\documentclass[manuscript,screen]{acmart}

\usepackage{float}
\usepackage[normalem]{ulem}
\usepackage{mathtools}

\AtBeginDocument{%
  \providecommand\BibTeX{{%
    \normalfont B\kern-0.5em{\scshape i\kern-0.25em b}\kern-0.8em\TeX}}}

\setcopyright{acmcopyright}
\copyrightyear{2018}
\acmYear{2018}
\acmDOI{XXXXXXX.XXXXXXX}

\begin{document}

\title[A Survey of Explainable Artificial Intelligence (XAI) in Financial Time Series Forecasting]{A Survey of Explainable Artificial Intelligence (XAI) in Financial Time Series Forecasting}

\author{Pierre-Daniel Arsenault}
\authornotemark[1]
\email{pierre-daniel.arsenault@usherbrooke.ca}
\orcid{https://orcid.org/0009-0005-8911-696X}
\author{Shengrui Wang}
\authornotemark[1]
\email{shengrui.wang@usherbrooke.ca}
\orcid{https://orcid.org/0000-0001-6863-7022}
\affiliation{%
  \institution{Université de Sherbrooke}
  \streetaddress{2500 Bd de l'Université}
  \city{Sherbrooke}
  \state{Québec}
  \country{Canada}
  \postcode{J1K 2R1}
}
\author{Jean-Marc Patenaude}
\email{jeanmarc@laplaceinsights.com}

\affiliation{%
  \institution{Laplace Insights}
  \streetaddress{230 King St. West suite 201}
  \city{Sherbrooke}
  \state{Québec}
  \country{Canada}
  \postcode{J1H 1P9}
}
\renewcommand{\shortauthors}{Arsenault, et al.}

\begin{abstract}

Artificial Intelligence (AI) models have reached a very significant level of accuracy. While their superior performance offers considerable benefits, their inherent complexity often decreases human trust, which slows their application in high-risk decision-making domains, such as finance. The field of eXplainable AI (XAI) seeks to bridge this gap, aiming to make AI models more understandable. This survey, focusing on published work from the past five years, categorizes XAI approaches that predict financial time series. 
In this paper, explainability and interpretability are distinguished, emphasizing the need to treat these concepts separately as they are not applied the same way in practice.
Through clear definitions, a rigorous taxonomy of XAI approaches, a complementary characterization, and examples of XAI's application in the finance industry, this paper provides a comprehensive view of XAI's current role in finance. It can also serve as a guide for selecting the most appropriate XAI approach for future applications.

\end{abstract}



\received{July 5, 2024}
\received[revised]{}
\received[accepted]{}

\maketitle

\section{Introduction}

In the last decade, Artificial Intelligence (AI) has made remarkable strides in terms of accuracy, transforming sectors such as business, media, healthcare, education, finance, scientific research, judicial, etc. While these advances offer unprecedented capabilities, they also introduce challenges, particularly in high-stakes, risk-sensitive domains like finance and healthcare. Despite their accuracy, AI models often suffer from a lack of user trust due to their black-box nature. Ethical and legal frameworks have evolved to reflect these concerns; reports such as \cite{oecd_artificial_2021}  and \cite{sarkar_explainable_2023} highlight the necessity for AI systems to be understandable in decision-making contexts. As pointed out by existing works \cite{mestikou_artificial_2023, oecd_artificial_2021, fsb_artificial_2017}, the poor interpretability of deep learning models can significantly increase investment risks, raising critical limitations for practical applications.

Addressing this problem is the emerging field of Explainable Artificial Intelligence (XAI), designed to produce AI models that are not only accurate but also understandable to human users. XAI offers two primary avenues for enhancing understanding: interpretable models, which are inherently understandable, and explainable models, which are models that require additional methods for explanation. However, there exists significant terminological confusion in the literature; the terms interpretable and explainable are often used interchangeably, leading to misunderstandings about the nature of the model being presented. Understanding the differences between these terms is critical for several reasons. Firstly, the pathways to achieve interpretability and explainability are different, necessitating distinct methodologies and evaluations. Secondly, accurate terminology helps in the selection of appropriate models based on the unique regulatory and ethical demands of financial applications. Thirdly, clear definitions improve interdisciplinary communication among data scientists, financial experts, and policymakers, facilitating more informed decisions. Fourthly, distinguishing between interpretability and explainability informs the level of trust one can place in a model's decision-making process, impacting risk assessment and compliance with financial regulations. Furthermore, the claim that a model is XAI sometimes lacks substantiation, either by failing to show the interpretations that could be done on the model or by lack of the explanations that help understand the model.

The need for this survey is further amplified by the financial industry's accelerating adoption of AI for complex tasks such as credit risk assessments, fraud detection, algorithmic trading, and, notably, the prediction of financial time series. Although these technologies promise to enhance efficiency and accuracy, they also raise significant concerns around fairness, explainability, and regulatory compliance. The opaque nature of AI algorithms is not merely a technological challenge; it has far-reaching implications for public trust, ethical governance, and regulatory adherence.
This paper aims to present the latest interpretable models and explainability methods used in the prediction of financial time series.
By offering a detailed survey of both interpretable and explainable models, it aims to serve as a guide for two groups of readers: data scientists who want to design and develop XAI into time series prediction, whether financial or not, and financial professionals who want to incorporate XAI into their analytics environment for predicting financial time series. Indeed, the interpretable and explainable approaches surveyed in this paper for predicting financial time series could also be used to forecast many types of time series. Also, this survey provides financial insights that the XAI made possible. Both the insights and their approaches can be useful for financial professionals. Additionally, this survey provides a rigorous foundation for continued investigation, guides practitioners in choosing the most appropriate approach, and informs them about the merits and limitations of various approaches. In essence, this survey aspires to facilitate the responsible and transparent application of AI in finance, aligning cutting-edge machine learning techniques with the sector's strict ethical and regulatory standards. To the best of our knowledge, this is the first survey focusing specifically on XAI for predicting financial time series.

In this paper, we categorize XAI models of the last five years that have been used to predict financial time series or to predict financial time series movement. Firstly, some important concepts are defined in the Section~\ref{definitions}. Secondly, the methodology of this survey and the diagram resulting from the taxonomy are presented (Sections \ref{methodology} and \ref{taxonomy}). Thirdly, interpretable models and explainability methods are presented and classified  according to its XAI principle (Sections~\ref{interpretable} and \ref{explainable}). In Section~\ref{alt_taxonomy}, the same approaches are characterized according to different criteria, namely the explanation characteristics known by the community \cite{das_opportunities_2020, kumar_overview_2023, banerjee_methods_2023}.
 Then, applications of the XAI in industry are presented (Section~\ref{applications}). Finally, conclusions and the work that remains to be done are presented (Section~\ref{conclusion}).

\section{Definitions}
\label{definitions}
The financial industry has increasingly adopted AI, since the inception of many statistical models in the 1980s, primarily for stock trading and risk management applications. As the domain matured, it embraced more advanced techniques, from expert systems and neural networks to the recent inclusion of deep learning models. These innovations have provided the industry with remarkable tools for market analysis, financial behaviour prediction, and risk management. However, the lack of transparency in these complex models often limits their practical application. When users cannot decipher a model's reasoning, even high-accuracy outputs might be deemed unreliable for critical financial decisions. This emphasizes the vital role of eXplainable Artificial Intelligence (XAI) as a solution.

The \textbf{eXplainable Artificial Intelligence (XAI)} is a branch of AI designed to be comprehensible to humans, thus promoting trust among users.
XAI's main feature is its ability to render the inner workings of its model intelligible to human users. Its importance in finance manifests in various ways. For instance, when a financial institution deploys AI for decision-making regarding a client, as in the case of a loan application, there is a pressing need to clarify the rationale behind such decisions. This transparency is imperative not only for individual clients but also for communicating decision-making processes to investors, board members, or auditors. XAI facilitates this clear communication. Additionally, when institutions rely on AI for critical decisions, stakeholders require insights into how those decisions are derived, ensuring their sustained trust in the system. From a risk management perspective, an opaque AI model introduces potential hazards. Comprehending the model's operations enables institutions to identify and mitigate risks more effectively. In decision-making scenarios, XAI provides pertinent information, helping users gauge the trustworthiness of predictions. Thus, users can judiciously combine the model outputs with their domain expertise to arrive at well-informed decisions. 

To qualify as XAI, an AI system must either be inherently interpretable (interpretable model) or use a distinct method that clarifies its decision-making processes (explainable model). An \textbf{interpretable model}, often referred to as a transparent model, is designed in such a way that users can intuitively understand its inner workings just by examining it. The individual components of the model are clear and understandable, facilitating a comprehensive grasp of its functionality. As an example, a linear model with a limited set of input features is considered interpretable. In such a model, each coefficient indicates the contribution of its associated feature. Thus, \textbf{interpretability} can be characterized as the inherent property of a model that, by its very design, makes its operations transparent at first sight. It is important to note that interpretability is not a simple binary trait. 
In \cite{lipton_mythos_2018}, the authors elaborated on the nuances of interpretability, highlighting dimensions such as \textit{simulatability} defined as the capacity of a model to be contemplated by a person,
\textit{decomposability} as the capacity of a model to be composed of interpretable parts,
and \textit{algorithmic transparency} as the level of interpretability of the model's training algorithm. A model is said interpretable if it is more understandable than an opaque model.

An \textbf{explainable model} is designed to provide explanations to users, aiding their comprehension of the model's operations. Often, an explainable model is essentially a black-box model enhanced with specific explainability methods, such as SHAP \cite{lundberg_unified_2017} or LIME \cite{ribeiro_why_2016}, to shed light on its internal processing. A pure \textbf{black-box model}, also known as an opaque model, operates in a manner that is not immediately discernible, making its internal workings and decision-making processes inscrutable to users. There are two reasons that some models are designated as black-boxes: either due to their inherent design or because of imposed confidentiality. For instance, deep neural networks, by virtue of their intricate architecture involving many operations, are inherently challenging to understand and thus are classified as black-box models. On the other hand, even a rudimentary linear model with a handful of features can be perceived as a black-box if the owner discloses only the outputs while withholding the coefficients. \textbf{Explainability} can thus be defined as the capability of a model to elucidate its functioning, offering users insights into its operational mechanics.

It is essential to define a few basic concepts that are frequently used in the rest of the paper. The term \textbf{model} refers to a series of calculations, or mathematical equations, describing the workings of a computing system that takes typical inputs, such as numbers or words, and generates outputs, or predictions, for a given task. 
An explainability \textbf{method} is defined as a process that takes a model as input and provides explanations of this model as output. Consequently, an explainable model is one on which an explainability method has been applied. An \textbf{approach} encompasses a number of these computation steps. We can define it as a process that takes an input and generates an output. As \cite{ochergykalo_fundamental_2023} explored, there remains ambiguity surrounding the fundamental concepts of XAI. It becomes essential to identify the target audience for the explanations and to delineate the nature of these explanations. Two core concepts in the preceding definitions appear nebulous:

\begin{enumerate}
    \item The term \textbf{user} is not explicitly defined. In the financial sector, this could refer to a data scientist, a business professional, an auditor, or even a consumer. Clearly, an explanation tailored for one may not be suitable for another. Consequently, it is crucial to specify the audience for whom the model should be rendered explainable or interpretable \cite{ochergykalo_fundamental_2023, bhatt_explainable_2020, barredo_arrieta_explainable_2020}.
    
    \item The notion of \textbf{explanations} lacks a formal characterization. These explanations must be context-sensitive \cite{belle_principles_2021, bhatt_explainable_2020}, addressing the distinct requirements of different users. For instance, in the realm of financial time series prediction, professionals often seek insights into feature importance to provide clarity on how input features influence the outcome. Another desired aspect of the explanation involves the decision rules, making the decision-making process clear and transparent. Time series interpretations also form a significant aspect of explanations, offering insights into the nature and patterns of input time series, and facilitating analyses based on extracted information and predictions. Furthermore, in the financial domain, visual representations such as user interfaces or dashboards are highly valued. Such visual aids not only present explanations but also offer supplementary information, fostering a more intuitive understanding of the model. Feature importance, decision rules, time series interpretation, and visual explanations constitute the \textbf{XAI principles} of the approaches  presented in the rest of the paper.
\end{enumerate}

 Let us apply these two definitions. For instance, if portfolio managers use the output of a model to assist in portfolio management, they are not seeking all types of interpretability or explainability. As they are not a data scientist, they may not necessarily aim to have a comprehensive range of explanations. Their specific goal is to determine whether to trust the model or to demystify if the model makes a mistake. To achieve this, a score that provides a confidence level of the model could be helpful. Additionally, the importance of features could also help them demystify if the model is making a well-founded decision or a poor one. 
They may not need the global importance of features, i.e., the importance of the features that contributed to the prediction of an entire dataset, but rather the local importance of features, i.e., the importance of features that contributed to a specific prediction.  
With local importance, they can understand how the model weighs its input information and why it made this particular decision. 
 On the other hand, the developer of the model looking to understand and improve his model may need a different explanation. What he might desire is the global importance of features, because the global explanations can be connected to the training of a model. By gaining a better understanding of the model, he could enhance it, for instance, by modifying the set of input features. In all cases, it's best to have both local and global explanations to obtain a better understanding of the model and its predictions. Moreover, if the features were produced by feature engineering or are simply features that have no financial sense, interpreting their importance might be challenging. Similarly, if there are too many features that take equal importance, it will not be useful for portfolio managers. A solution to this problem could be to group features into financial themes (e.g., volatility, momentum, periods, etc.) that would be understandable for the portfolio managers. 

The previous discussions suggest that the explanations should be adapted to the context and the user for optimal effectiveness. However, the evaluation of the explanations should not be subjective. It should be based on rigorous criteria. Adjustment to the context could be one of these criteria, but the evaluation itself should not depend on the subjectivity of the user. We present some ideas of criteria in Section~\ref{explain_eval}.

\section{Methodology}
\label{methodology}
In this paper, we categorize approaches  that are either explainable or interpretable and used for predicting financial time series or their movement. These time series may include stock prices, volatility, and macroeconomic indicators, among others. We conducted a comprehensive search in various databases, including \href{https://www.scopus.com/search/form.uri?display=basic}{SCOPUS}, \href{https://www.engineeringvillage.com/search/quick.url}{Inspec}, \href{https://ieeexplore.ieee.org/Xplore/home.jsp}{IEEE Xplore}, \href{https://web.p.ebscohost.com/ehost/search/advanced?vid=0&sid=6ccde257-39aa-465b-97b3-cee405cc2da0\%40redis}{Computers and Applied Sciences Complete}, \href{https://dl.acm.org/}{ACM Digital Library}, and \href{https://arXiv.org}{arXiv}. Our search criteria included articles containing the following keywords: "explainability", "finance", "forecasting", and "model". Alternative terms for "explainability" could be "interprétabilité", "interpretability", "XAI", "explainable", "interpretable", "explicability", "feature importance", or "transparent model". For "finance", substitutes include "stocks", "volatility", "tendency", "financial", and "market prices". Similarly, "model" could be replaced by "AI", "method", "machine learning", or "artificial intelligence". We limited our search to papers published between 2018 and June 2023.

We applied two filters to the papers we found. First, we read the titles and abstracts, retaining papers that focused on finance and XAI. Examples of papers that we excluded include those about XAI in ecology, papers that were finance-focused but did not incorporate XAI, or papers on finance that did not employ machine learning. Second, we reviewed the essential sections of the shortlisted papers to extract the necessary information related to XAI. If a paper focused on the prediction of financial time series or their trends using an explainable or interpretable model, we kept, understood and summarized its XAI part. Otherwise, we excluded the paper from our review. For example, we eliminated papers that presented models forecasting credit risk and loan approvals because these models did not predict a time series or its trend. Indeed, they predicted a number or a class that did not depend on the time. 
Our emphasis is on the interpretability and explainability aspects of the models discussed. The aim of this paper is to provide an overview of advancements made in the explainability and interpretability of models designed for predicting financial time series. Some models use time series data as input, others rely on text from news articles or the internet, and some utilize both. 

It is true that some articles received more attention than others in this review; however, this should not be interpreted as an indication of their relative importance. We may also have inadvertently omitted articles that merit being included. Indeed, our review process involved reading specific sections of the papers to focus on the interpretability and explainability of the models. While our aim was to conduct as a thorough search as possible, there is a chance that we may have missed some relevant articles on the topic.

\section{Taxonomy}
\label{taxonomy}
\begin{figure}

    \makebox[\textwidth][c]{\includegraphics[width=0.95\textwidth, keepaspectratio]{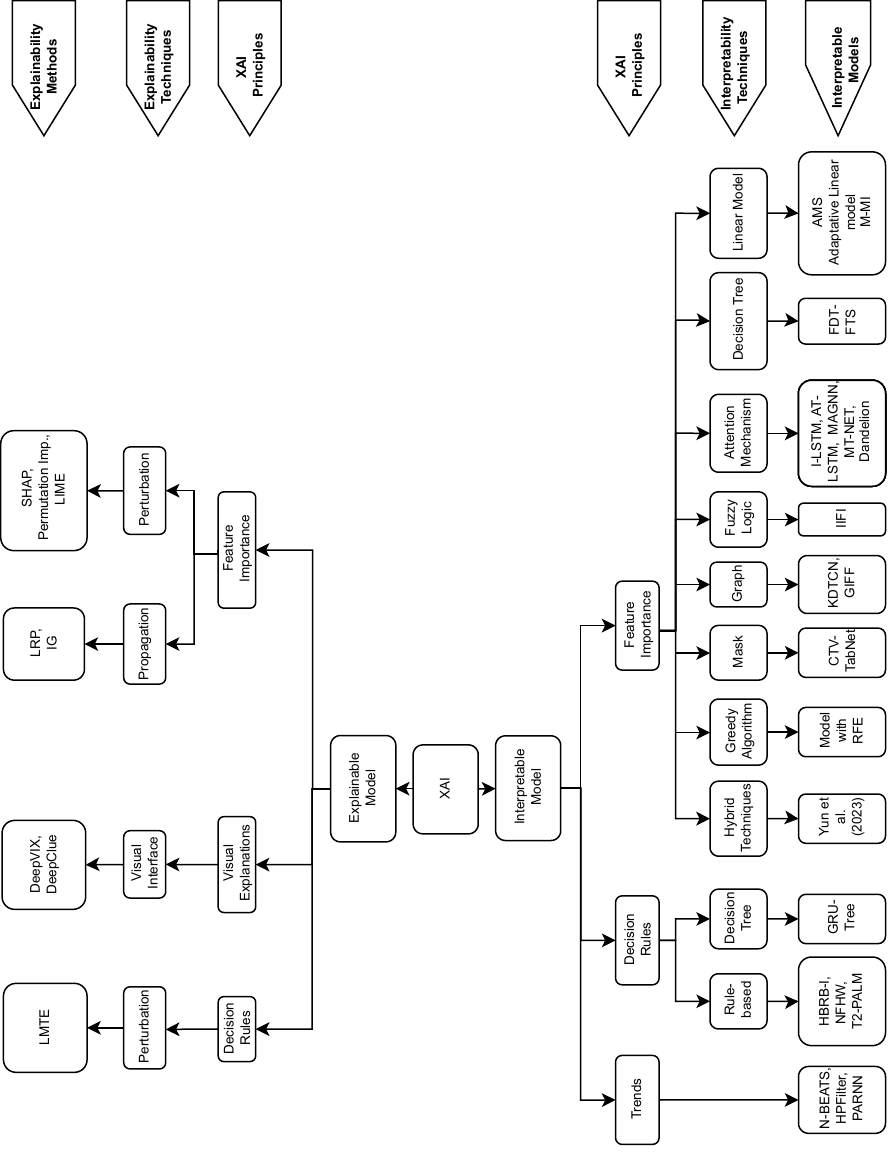}}
    \caption{Classification of XAI Approaches }
    \Description{This diagram illustrates the XAI taxonomy proposed in the paper, encompassing four levels of classification. Initially, XAI is categorized into Explainability Models and Interpretable Models. Subsequently, both are further classified based on XAI principles. Following this, they are classified according to the technique. Lastly, the names of specific models and methods are provided within the respective categories.}
    \label{fig:classification_methode}

\end{figure}

The goal of the taxonomy presented in the Figure \ref{fig:classification_methode} is to help readers identify the right XAI approach for their specific context. Initially, XAI approaches are divided into interpretable models and explainable methods. Each category is then further sorted based on the principles of XAI, including feature importance, visual explanations, and time series analysis. After that, explainability methods and interpretable models are categorized based on the technique used to illustrate the XAI principles. For the interpretable models, it includes linear regressions, attention mechanisms, decision trees, graphs, etc. For explanation methods, it includes perturbation, propagation and visual interface. The reviewed approaches are presented in the Sections \ref{interpretable} and \ref{explainable} following the diagram presented in the Figure \ref{fig:classification_methode}.

\section{Interpretable Models}
\label{interpretable}
 
This section presents the interpretable models reviewed in the survey. These models provide feature importance (see Subsection~\ref{sec:feature_importance}), decision rules (refer to Subsection~\ref{decision_rules}), or analysis of relations like trends in the time series (as discussed in Subsection~\ref{TS_interpretation}). The evaluation of the models' interpretability is discussed in Subsection \ref{inter_eval}.

Some of the presented models may appear to be more like black-box models than interpretable models. These models, because of their complexity, do not fit the conventional definition of interpretable models that only regroups basic models. For instance, neural networks, attention models and other complex models are usually considered as black-box models by the community. This makes sense because they are complex compared to basic interpretable models such as linear models and decision trees. Based on that criterion, complex models, like the attention models, would
have been excluded from the class of interpretable models.
However, the XAI models are not classified according to their complexity. To be considered as interpretable, the models need to provide some information that helps the user understand them, specifically through XAI principles. Therefore, the origin of the XAI principles is the key factor that determines the nature of a model with regard to interpretability and explainability. If the principle directly comes from the model itself, the model is considered interpretable. Otherwise, if the XAI principle comes from another algorithm, namely an explainable method, the model will be considered as an explainable model. So, even if some models are complex, they can still generate information that can be interpreted by the user to better understand the process behind the model. This is why many authors, including us, refer to them as interpretable. For example, attention models, like the ones presented in Subsection \ref{attention}, contain attention weights that represent the importance of the features. As these weights come from the models themselves, they are classified as interpretable models. It is also the case with complex neural networks that are classified as interpretable due to a component in their structure, like a mask \cite{seo_exploring_2023} or components of fuzzy logic \cite{wang_interpretable_2023}, that reveals some information about the networks to users.

There may exist other definitions of interpretability and explainability that would lead to a different classification of those complex models. We believe that our definitions in this paper help draw a clear line between interpretability and explainability. And the classification of those models according to these definitions is logical and easy to understand.

\subsection{Feature and Time Importance}
\label{sec:feature_importance}

\textbf{Feature importance} is an XAI principle that measures how much each feature has an effect on the prediction or on an entire dataset. It also includes time importance, which measures how much specific time periods were important.
If the feature importance reflects the effect on a specific prediction, it is called local feature importance. If it reflects the effects on a dataset, it is called global feature importance. Usually, the global feature importance is computed on the training dataset, but it can also be computed on the test set. By ranking the features according to their importance, the important features can be extracted.
The feature importance can be plotted through heat maps, bar plot, line plot, etc. The feature importance is computed for all types of features, including both input features and internal features, which represent learned representations within the model. It is important in finance because it shows which information the model uses to make its predictions. The comparison between financial analysts and a model can easily be made. In their work, the analysts will, unconsciously or consciously, weight the information that they will read and make a prediction based on the important one. A model, showing the importance of its features, will give the same kind of information to the user. 

The feature importance is connected to features themselves. Indeed, having a high number of features can obscure the understanding of each feature's importance. To bypass this issue without altering the model itself, one strategy is to aggregate features into clusters and then assess their importance at the cluster level. If some features do not have a financial meaning, it becomes more challenging to interpret the results. When they do, it is possible to draw more detailed conclusions about interpretability. This connection between the model and the context enables users to better understand both. When the context is well known, feature importance can help justify the model's decisions. Furthermore, the feature importance of an accurate model can reveal unknown information about the financial market and help users develop new theories. In this paper, various techniques to compute feature importance are discussed. These include coefficients in a linear model, gain in decision trees, relevance scores, and attention weights. It is important to note that while all these techniques aim to measure feature importance, they might not all convey the same underlying concept. Therefore, it will be important to evaluate the differences between these approaches in the future.

\subsubsection{Linear Regressions}
\label{lin_reg}
The first model that comes to mind when discussing feature importance is the linear model. It is one of the most interpretable models due to its simple way of learning and the feature importance that it provides. However, a vanilla linear model cannot precisely predict financial time series due to their complexity and non-linearity. For this reason, the vanilla linear model is not used in practice for predicting financial time series. However, linear regression can be integrated into a complex model that has the ability to predict financial time series, enhancing its interpretability.
We define the linear regression $f(x)$ as $f(x) = \beta_0 + \sum_{j} \beta_j x_j$, where $x_j$ are the features and $\beta_j$ are the coefficients of the associated feature $x_j$. These coefficients represent the effect of their associated features. The features can be the input ones \cite{xu_adaptive_2020, munkhdalai_recurrent_2022}, but they can also be internal features \cite{zhang_stock_2018}.

\begin{figure}[!ht]
\begin{center}
    \includegraphics[width=0.8\textwidth, keepaspectratio]{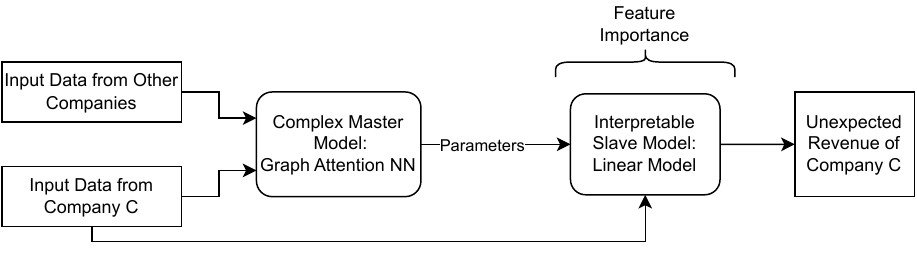}
    \caption{Architecture of AMS}
    \Description{Input data from company C is directed to both the interpretable slave model, represented by the linear model, and the master model, which is the graph attention neural network. The master model provides parameters to the slave model. The output of the slave model is the unexpected revenue of company C. Feature importance is derived from the slave model.}
    \label{fig:ams}
\end{center}
\end{figure}

A common way to use a linear regression in a complex model is to compute the weights of the regression with a complex model. This way allows a trade-off between interpretability and accuracy: it loses in the simplicity of the algorithm to learn the weights, but it gains in accuracy. For example, in order to predict unexpected revenues for companies, an Adaptive Master-Slave (AMS) model was proposed by \cite{xu_adaptive_2020}. The model has different weights for each company, unlike the normal linear models that would have the same weights for all companies. Thus, each company has their important and non-important features that help the user to understand more locally the predictions. As its name suggests, the model is composed of a master model and a slave model. The master model, a graph neural network, creates a slave model for each company. The slave model is a linear regression. Subsequently, the slave model predicts the unexpected revenue for each company using the input features. Since the slave model is a linear one, it is possible to extract the importance of features for each company. A flow chart of AMS can be seen in the Figure~\ref{fig:ams}. \citeauthor{munkhdalai_recurrent_2022} [\citeyear{munkhdalai_recurrent_2022}] proposed also an architecture composed of a linear model and a complex model. In that case, the complex model is a recurrent neural network. Indeed, to predict time series, including the price of the Nasdaq, from temporal numerical data, the authors \cite{munkhdalai_recurrent_2022} use an adaptive linear regression. It is adaptive in the sense that its coefficients change over time. Initially, a linear regression is learned throughout the training. Then, the recurrent network determines the changes to be made to the coefficients in order to adapt them through the time. Finally, the weights are updated. The prediction model is therefore interpretable locally by the coefficients of the linear regression. The average of the coefficients were done to compute the global importance. 
For example, the global importance of the "DTWEXB\_1", that is 1 day lag of trade weighted U.S. dollar index is -0.58. It means that an increase of one point of this feature would cause a drop of a 0.58 points of the Nasdaq index\cite{munkhdalai_recurrent_2022}.
Since the coefficients of the model change through times, they were plotted as time series. These time series can be analyzed with classic statistics like the trend, the volatility, etc. For instance, it shows that the importance of "DTWEXB\_1" was highly volatile between 2015 and 2016. It also shows that "DTWEXB\_1" was  more important in 2017 than before. 
This model does not show the importance of the features, but of the partial derivative of the features.

A different way of integrating linear regression into a complex model involves projecting the input into meaningful dimensions and then applying a linear combination of these dimensions with the appropriate weights. This is the concept behind the semantic path model described by \cite{feuerriegel_news-based_2019}, where text is used as the input. Indeed, this model helped the authors to better understand the economic context of the Europe. More specifically, it projects the input words into a reduced latent semantic space and forecasts macroeconomic indicators. The dimensions of the latent space are interpretable and are represented by types of words such as "negative outlook", "economic uncertainty", "legal risk", and so on. These types are then linearly summed to maintain interpretability and produce the final prediction. The importance of the dimensions of the latent space was analyzed and some financial conclusion has been made. For example, the authors found that the negative dimension, that represent negative news, affects many indicators, namely the unemployment rate, inflation, industrial production, consumer confidence, and government bonds. 
The authors demonstrated that the way the model is constructed enables users to make connections between macroeconomic indicators and input words through the bias of types of words. A similar model using linear regression and text as input was proposed in \cite{zhang_stock_2018}.

\subsubsection{Decision Trees}
Decision Trees are among the most interpretable machine learning algorithms. A decision tree models decision making through a series of questions based on feature values, leading to a clear, tree-like structure that can be easily visualized and understood. Each node in the tree represents a decision criterion, and each subsequent branch represents an outcome of that decision. The final leaf nodes provide the model's prediction. Their transparent nature is reflected in that the reasoning behind any specific prediction can be traced back through the tree, making it straightforward to explain why a particular decision is made. This clarity in decision-making processes makes decision trees the candidates of choice in scenarios where interpretability is crucial. Usually, decision trees are interpretable by the rules they provide to understand the decision process (Subsection \ref{decision trees}). However, decision trees can also be interpretable by providing the gain of each feature. The gain measures a continuum of feature importance from global to local. The gain represents how much the node of a feature contributes to decreasing the uncertainty of the model during the training.

As an example of applying decision trees to financial time series forecasting, \citeauthor{silva_c45_2021} [\citeyear{silva_c45_2021}] used decision trees to predict the closing price of IBOVESPA with historical values as input and technical analysis. More specifically, they introduced the FDT-FTS (Fuzzy Decision Tree-Fuzzy Time Series) method for this task. As the name suggests, this approach consists of a fuzzy decision tree induced from the C4.5 algorithm \cite{quinlan_c45_1993}. The method has four parts: data analysis, data fuzzification, training and testing of the FDT, and results defuzzification. 
Also, the importance of the features is computed from the information gained during tree induction. According to the gain, the most important features are the RSI, the difference of moving averages and the first difference delayed in t-1.

\subsubsection{Attention Mechanism}
\label{attention}
Attention mechanisms in deep learning models, such as the conventional attention and the self-attention found in Transformers \cite{vaswani_attention_2017}, offer interpretability by assessing the importance of input features \cite{guo_exploring_2019, zhang_at-lstm_2019, yang_explainable_2019}, internal features \cite{lin_kernel-based_2022, cheng_financial_2022, zhou_domain_2020}, or specific time periods \cite{guo_exploring_2019, chang_memory-network_2018, tran_temporal_2019}. Indeed, the attention mechanism is based on correctly weighting the inputs to make a better prediction. It provides an easy way to include the time importance in the interpretation.

In fact, the time importance is desirable for a financial analyst to understand which time frame the model utilizes for its predictions. The time frame information helps him/her determine if there was a specific event that happened during that time. This event could be a significant decrease in a stock's price or a company sale. Such information not only helps the analyst figure out if the model's prediction could be incorrect, but also provides a deeper understanding of the context. Attention models offer this kind of insight through their two-dimensional attention weights. Since attention models take sequential data with both a time dimension and a feature dimension as input, the attention weights also have these two dimensions. In fact, these weights indicate where the model is focusing its attention, on which features and at which time points. 
The weights can then be extracted for model interpretation and are often visualized through heatmaps where one dimension represents time and the other represents features. This offers insights into the model's selective focus and its consideration of contextual relevance. To obtain the absolute time or feature importance, the weights can be summed according to the right dimension. We present some attention models that can be used to analyse feature and time importance of sequential input such as time series, words and events.

  The two first attention models \cite{liu_improving_2022, zhang_at-lstm_2019} described here contain an LSTM integrated with an attention mechanism.
  \citeauthor{liu_improving_2022} [\citeyear{liu_improving_2022}] used the ILSTM (Interpretable-LSTM) from \cite{guo_exploring_2019}. This LSTM is unique because of its ability to separate hidden states by features, making it possible to discern their individual effects on the prediction. The model contains a mixture of attention mechanisms, namely temporal and variable wise attention, which consolidates these hidden states for forecasting and provides feature importance. The task of \cite{liu_improving_2022} was to predict CO2 flow based on several meteorological time series. These series exhibit seasonal variations. The model demonstrates which series and at what points in time they were important for the prediction. 
  It also shows the importance of features over time during training. \citeauthor{liu_improving_2022} [\citeyear{liu_improving_2022}] indicated that beyond helping understand the model, the interpretations assist understanding of the context of the problem and increasing of the performance of predictions. The authors of ILSTM \cite{guo_exploring_2019} had tested this model on financial data in an experimental work, and the findings were promising. The task was to predict the NASDAQ 100 based on companies indexed beneath it. Their model outperformed other opaque models while also being interpretable. It shows the importance of each company at the different time step lags and at different epochs. The three most important stocks to predict the NASDAQ 100 are NTAP, FOXA and TRIP according to their model.

\begin{figure}[!ht]
\begin{center}
    \includegraphics[width=0.8\textwidth,keepaspectratio]{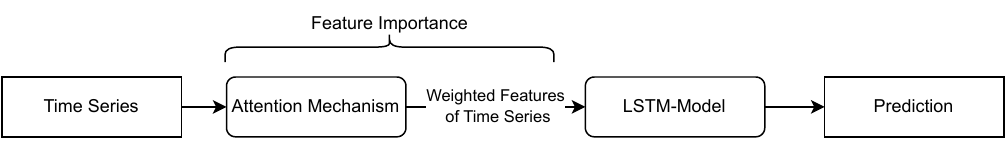}
    \caption{Architecture of AT-LSTM} 
    \Description{This is a flow chart of the AT-LSTM model from \cite{zhang_at-lstm_2019}. The time series serves as the input, passing through the attention mechanism, which outputs the weighted features of the time series. These features are then fed into the LSTM model to make predictions.}
    \label{fig:atlstm}
\end{center}
\end{figure}

Similarly, a distinct adaptation of LSTM was presented in
\cite{zhang_at-lstm_2019}. The model is named attention-based LSTM (AT-LSTM) and was used to predict stock prices. The unique part of this model is its attention module placed before the LSTM. This module is used to weight the input features. The basic idea that the model relies on is that input features should not all have the same weight (as in a standard LSTM). Like a financial analyst 
would do, the model needs to weigh the information it receives to make a good prediction. In addition to improving the LSTM's performance, this module adds a layer of interpretability. Indeed, the user can visualize the importance of features from the attention weights. The architecture of the model is presented in the Figure~\ref{fig:atlstm}.

 The Multi-modality graph neural network (MAGNN) was presented in \cite{cheng_financial_2022}. MAGNN was used to predict financial time series. Unlike typical regression models, MAGNN uses not only time series as input but also events and news. These types of input are named "modalities" by the authors. The model mainly consists of attention mechanisms and graphs. There are two types of attention in the model: inter-modality attention and inner-modality attention. The inter-modality attention is used to determine the importance of each modality depending on the different market situations, i.e., the different predicted returns. For example, the inter-modality attention proved that all the modalities are important in the prediction. Also, it showed that the event modality seems to be more relevant in high-return situations, unlike the market and equity modalities, which are more important in low-return situations. It can be explained by the fact that news have a significant influence on sudden equity changes. 
 It makes sense because the news explicitly expose important stories like acquisitions of companies. The inner-modality attention, slightly less interpretable, is used to learn the relations between input sources and the target prediction. This relation is represented in terms of a graph. The number of neighbours of a node in a subgraph is used as a variable to analyze the feature importance. 

Again for financial time series prediction, the Domain Adaptive Multi-Modality Neural Attention Network, named Dandelion, was proposed in \cite{zhou_domain_2020},  and was designed to be interpretable by end users while delivering good results in predicting company revenues. This model is unique due to three components. Firstly, it exploits the interconnections between various modalities. The modalities were, in that paper, weather, finance, and news. 
Secondly, the model has the capability to  "capture the high-level domain knowledge of stocks and understand the underlying importance and logic of modalities/variables across different domains." \cite{zhou_domain_2020}
Finally, it explains outputs by revealing feature importance, showing relevant domain and time stamp details. The model's trinity attention indicates the importance of features based on the task, modality, and time. For example, it indicated  that the temporal importance increases over times. It suggests that recent information is more important than distant information, which makes sense. Also, it revealed that finance features are the most important among the weather, news, Wall Street and web features.

While attention weights provide a glimpse into a model's processing, it is not always a definitive representation of its reasoning, and its interpretability may vary across tasks. This will be discussed in the Subsection \ref{inter_eval}. 

\subsubsection{Fuzzy Logic}

The model presented in \cite{wang_interpretable_2023} makes use of fuzzy logic to show the feature importance while, typically, most existing models such as \cite{hassanniakalager_conditional_2020, xie_embedded_2022, xie_interpretable_2021, rajab_interpretable_2019, mingxiang_guo_deep_2021, ferdaus_multiobjective_2022, cao_multiobjective_2020} make use of it to compute decision rules. The authors of \cite{wang_interpretable_2023} proposed an architecture that employs \textit{Intuitionistic} fuzzy logic and deep learning for stock prediction tasks. Unlike the general fuzzy logic that employs a single value between 0 and 1 to represent membership, intuitionistic fuzzy logic employs three, i.e. membership, hesitation, and non-membership. These concepts can be interpreted respectively as the degree of support for the rise of stocks for the membership value, the degree of rejection for the rise of stocks for the non-membership (non-affiliation) value and the  uncertainty of the prediction for the hesitation value according to \cite{wang_interpretable_2023}. Feature importance was measured by using the hesitation concept. The features were masked one by one for all stocks. If the hesitation increases when a feature is masked, it indicates that the feature has an effect on a prediction. The average magnitude of the change in hesitation estimates the feature importance, allowing significant features to be identified. The impact of features on rising stocks was analyzed with the membership values. To do so, the features were masked one by one, but only for rising stocks. When a feature is masked, an increase in the membership value suggests a higher probability that the predicted rising stock will actually rise. The magnitude of the change shows the contribution of the masked feature to stocks' rise.  The architecture of the model enables a better understanding of the model's predictions. The proposed structure is named Interpretable Intuitionistic Fuzzy Inference Model (IIFI). The IIFI has six layers: the input layer, the fuzzification layer, the encoding layer, the interpretation and inference layer, the consequent layer and the defuzzification layer. 

\subsubsection{Graph}

Often, when graphs are incorporated within a model, they allow capturing relationships among features and estimating the importance of these features. This makes the model interpretable, as graphs are generally easy for humans to understand. 

Using a textual network, \citeauthor{li_incorporating_2020} [\citeyear{li_incorporating_2020}] analyzed the relationship between a stock index and analysts' research reports. Although, it is not directly related to time series prediction, their process predicts the stock movement using a sparse Laplacian logistic regression. This logistic regression is based on a graph that is constructed from analysts' research reports. Specifically, 56 most representative words,  termed as "representative textual features", were extracted from the reports and used as the graph's nodes. An edge between two words was established if the two words were statistically frequent in the same window in a report. Subsequently, the graph was interpreted via the most influential words based on their centrality, the graph's density, closely interlinked subgroups, etc. The importance of each word is presented in the article. "Imagine", "Pessimism" and "Effectiveness" are the words having the highest positive impact, while "Concern", "Design" and "Chengyu stock" have the highest negative impact. The negative words seem to have a higher impact than the positive words according to their importance coefficient.

Another interpretable model using graph was presented in \cite{deng_knowledge-driven_2019}. It is named Knowledge-Driven Temporal Convolutional Network (KDTCN). The model takes stock price values and news as input, and predicts stock movements. The architecture is divided into two modules: the Knowledge-Driven event embedding (KD), followed by the Temporal Convolutional Network (TCN). The KD module creates a knowledge graph based on events in the news, and returns the "event embeddings". Then, the TCN module takes these embedded events concatenated with the prices as input to compute the prediction. The architecture is interpretable because of the effect of events on the prediction and the relationships between events that can be analyzed through graphs, like \cite{li_incorporating_2020}. Indeed, within their TCN framework, the effects of different events are easily extracted from the TCN, therefore significant events are identified in the graphs created in the KD module. The direct and indirect links between events can thus be analyzed, providing a deeper understanding of the decision-making context. For example, "[United Kingdom, votes to leave European Union]" has a high effect on the prediction of the stock to rise, according to \cite{deng_knowledge-driven_2019}.

Yet, to predict stock trends, \citeauthor{wang_interpretable_2020} [\citeyear{wang_interpretable_2020}] developed the Graph-Based Interpretable Stock Trend Forecasting Framework (GIFF). GIFF is divided into three modules: the sequential embedding module, the graph residual module and the forecasting module. A distinctive feature of this architecture lies in the graph residual module. This module is subdivided into three layers: the business-based layer, the industry-based layer and the fully connected layer. In the first and second layers, connections are established between stocks that share the same businesses and industries, respectively, forming graphs. The last layer connects every pair of stocks. Since it is partitioned in this manner, the layers are interpretable and the results can be interpreted for each different industry. For instance, the authors compared the correlation between the predicted returns and the actual returns of the same industry (e.g., electronics). This allowed them to determine how well their model predicts for each economic sector and individual business. However, compared to \cite{li_incorporating_2020}, this approach does not allow the direct measurement of the feature importance.

Similarly, the model proposed in \cite{huang_ml-gatmultilevel_2022} does not explicitly compute the feature importance. Instead, it highlights relations between the inputs. This model, named Multilevel Graph Attention Network (ML-GAT), predicts stock movements by taking as input numerical  from Yahoo Finance website, textual from news, and relational data from Wikidata. Essentially, the model extracts features from various types of input: it encodes text-data and price data, and derives graphs from company relationships. Subsequently, the text and price data are incorporated into the graphs. Ultimately, the price trend is forecast. This model is interpretable due to the insight that can be drawn from the created graph. The relationships reflect reality since they are derived from real-world data, allowing users to obtain a comprehensive visualization of the situation. For example, a second-order relation ($A \xrightarrow[]{R_1} B \xleftarrow[]{R_2} C $, where $A, B, C$ are entities of Wikidata) could be defined as: $R_1$ represents the "owner of the subject" and $R_2$ represents "object of which the subject is a part" \cite{huang_ml-gatmultilevel_2022}. So, $B$ would be the owner of $A$ and $B$ is an object of which $C$ is a part.

\subsubsection{Mask}

\begin{figure}[!ht]
\begin{center}
    \includegraphics[width=\textwidth,keepaspectratio]{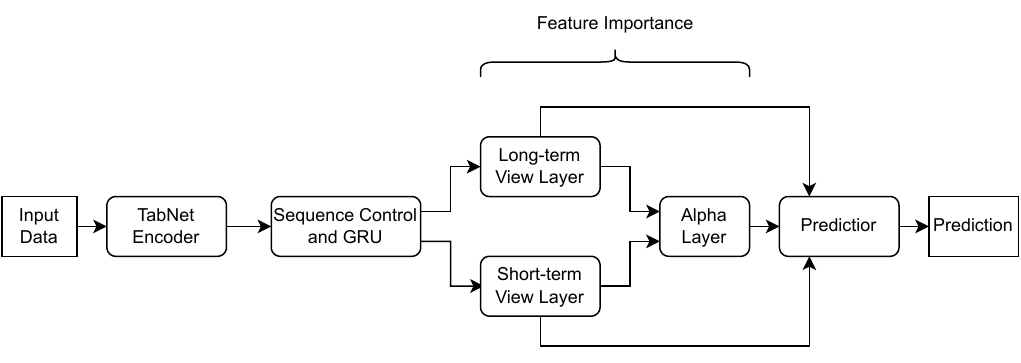}
    \caption{Architecture of CTV-TabNet}
    \Description{The input data is first fed into the TabNet encoder. The output of the encoder is then directed to both the sequence control and GRU components. The outputs from these components are further directed to two layers: the short-term view layer and the long-term view layer. The outputs from these layers are subsequently passed to both the alpha layer and the predictor.}
    \label{fig:ctvtabnet}
\end{center}
\end{figure}

The Combined Time-View TabNet (CTV-TabNet) proposed in \cite{seo_exploring_2023} performs a stock movement prediction task with  candlesticks and moving averages as input. This deep architecture consists of three parts: the TabNet encoder, the Sequence Control and GRU, and the Multi-Time Views Combiner. Its interpretability comes from the feature importance provided by the encoder's masks and the alpha layer of the Combiner, which determines the importance of both short-term and long-term views. The importance of the features was extracted and analyzed. A flow chart of the architecture is presented in Figure~\ref{fig:ctvtabnet}.

\subsubsection{Greedy Algorithm}
To classify intraday stock movements, \citeauthor{kumar_blended_2021} [\citeyear{kumar_blended_2021}] proposed an architecture that includes a module for calculating feature importance. The classifier's architecture is composed of three modules: recursive feature elimination, feature importance, and deep neural network. Feature importance is calculated following a recursive feature elimination phase, prior to the processing by the deep neural network. The least significant features are removed using the greedy recursive feature elimination (RFE) algorithm. In short, this algorithm calculates the effect of eliminating subsets of features on accuracy to retain the subset of features that maximizes the accuracy. Thus, feature importance is extracted directly from the effect of the features on accuracy and can be analyzed. Thanks to this feature elimination process, it is categorized as an interpretable architecture.

\subsubsection{Hybrid Techniques}

 Several models make use of a mix of techniques to provide feature importance.  For instance, the approach presented in \cite{yun_interpretable_2023} uses tree-based models (XGBoost, Random Forest, Decision Tree, Light Gradient Boosting Regression, and ExtraTrees), a two-stage feature selection process, and a local interpretability engineering module to compute global feature importance, local feature importance, and time importance.

The focus of the approach in \cite{yun_interpretable_2023} is on increasing the accuracy and the interpretability of raw tree-based models.
Indeed, the importance scores from tree-based models are sensitive to feature permutation and to the training conditions. To compensate for this limitation, a genetic algorithm is used in the first phase of the two-stage feature selection module to test and find the best combination of features for each tree-based method. In this phase, the global feature importance comes from the decision tree model. In the second phase, the features with an importance score higher than a cutoff are retained to form the optimal feature subset. To add local interpretability to their model, the authors convert this global feature importance to local feature importance and time importance. This conversion is achieved using a curve fitting algorithm. Then, both global and local forecasting are performed. The two-stage feature selection enhances the robustness and optimality of the feature subsets and, consequently, the estimation of feature importance. For this reason, it is classified as an interpretable model.

\begin{figure}[!ht]
\begin{center}
    \includegraphics[width=0.8\textwidth,keepaspectratio]{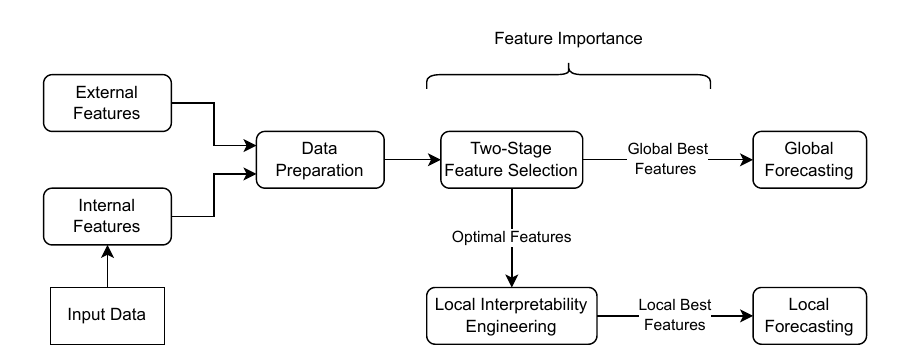}
    \caption{Architecture of the Model  in \cite{yun_interpretable_2023} }
    \label{fig:feature_sel}
    \Description{The input data is directed to the internal feature module, and both the internal and external features modules contribute to the data preparation phase. The output of this preparation is then passed to the two-stage feature selection module. The global best feature sets derived from this module is directed to the global forecasting component. Simultaneously, the optimal feature sets are forwarded to the local interpretability engineering section. This part outputs the local best features, which are then fed into the local forecasting process.
    }
\end{center}
\end{figure}

The task of  \cite{yun_interpretable_2023} was to predict the XOM price using both internal features which are technical indicators of XOM, and  external features which are time series not derived from XOM. 
According to the selection method, the most important internal features were the weighted closing price and the balance of power. Indeed, the transformed prices features are all important because these features are based on past prices, which have an effect on future prices, according to \cite{yun_interpretable_2023} and \cite{ostrom_time_1990}.  
Based on research by \cite{boxer_profitable_2014}, the authors indicated that the balance of power "implies the direction of a trend measuring the strength of buying against selling by assessing the ability of each side to drive prices to an extreme level" \cite{yun_interpretable_2023}, which would explain its high importance.
Also, the external features analysis revealed that both SPE and VLO were important to the prediction for all data segments. They differed slightly when it comes to the local feature importance. It was the SPE that is the most important feature for three of the four time segments. This study shows that internal and external features have predictive power. However, the internal features seem to be more effective than the external ones. The architecture of the approach is presented in the Figure~\ref{fig:feature_sel}.

 Another model, proposed in 
\cite{zheng_incorporating_2021}, uses a mix of regression technique and financial theory to provide interpretation.  Indeed, it consists of several individual neural networks and one weighting neural network. The latter determines the voting on the individual networks, each of which predicts separately the target value which is the implied volatility. Therefore, the voting weights represent the importance of each neural network. This technique is similar to the one presented in section \ref{lin_reg}. Additionally, financial theories are incorporated into the architecture and the training process to align the model with financial theories. For example,  the "smile function" \cite{zheng_incorporating_2021} was added to the model as an activation function.

\subsection{Decision Rules} 
\label{decision_rules}

Decision rules are another XAI principle that contributes to the interpretability of a model. They prescribe the conditions a model employs to make predictions, reflecting the patterns learned from the data. In some cases, these rules can be linked to financial theory \cite{wu_preliminary_2019}. By revealing these rules, users develop a clearer understanding of the model's operations and can find patterns that they might not otherwise guess.

There are two main techniques for providing decision rules: from a decision tree or from a rule-based module. Even though the decision tree is a type of model, it can be used within a complex model and retains its interpretable nature. On the other hand, a rule-based module is a component of a model, or the entire model itself that contains the decision rules. It includes all specific methods for computing the rules such as rules derived from experts, learned rules, and a mixture of both. By convention, a rule-based module does not include decision trees.

\subsubsection{Decision Trees}
\label{decision trees}
Decision trees is a technique that provides decision rules. Each decision tree's node represents a decision made by the model. The decision trees differ from other rule-based modules in the algorithm for trees construction. Such an algorithm often utilizes information gain from features to construct the trees. This gain can also be extracted to provide feature importance. According to \cite{carta_explainable_2021}, decision trees must be relatively short to maintain a low number of rules and, consequently, be interpretable.

A way to use decision trees is to transform a complex model into a decision tree and then interpret the decision tree. Indeed, to predict stock price movements, the approach introduced in \cite{wu_preliminary_2019} uses a GRU and a GRU-Tree. The GRU-Tree is a GRU to which the regularized tree algorithm proposed by \cite{wu_beyond_2017} has been applied. This algorithm transforms a GRU into a decision tree. Thus, the GRU-Tree is intrinsically interpretable for the same reasons as a normal decision tree. The process is in three steps. First, the stock price is predicted with the GRU. Second, the regularized tree algorithm is applied and a decision tree, i.e. the GRU-Tree, is obtained from the GRU. Third, the GRU-Tree is interpreted. One interesting thing is that the authors found that some branches of the tree exposed the mean reversal, a known rule in finance. They also used the GRU-Tree to predict the stock price movements and computed the fidelity between the GRU and the GRU-Tree to be about 0.8. Therefore, GRU-Tree acts as an 
 interpretable proxy model of GRU, allowing the predictions made by the GRU to be understood and analysed in the form of a decision tree.

To upgrade their model, the same group of authors improved their GRU-Tree with a L1-Orthogonal Regularization to form the L1-Orthogonal Regularized GRU Decision Tree \cite{wu_stock_2020}. The L1-Orthogonal Regularization is incorporated in  GRU's loss function. This regularization makes the GRU more amenable to approximation by a decision tree. Specifically, the L1-Orthogonal regularization term reduces the number of parallel decision boundaries determined by the GRU, simplifying its representation for a decision tree. This regularization forces the model weights to be orthogonal. Once the GRU is optimally trained, the decision tree is trained using the features from the training set as input and the values predicted by the GRU as both the target and predicted values. After training, the decision tree is interpreted, and significant rules are extracted. Simple moving average and weighted moving average features were predominant in the rules predicting "down" or "up". The fidelity measure  was shown to have been improved to around 0.97, thanks to the L1-Orthogonal regularization term.

Another, and simpler, way to use a decision tree is to incorporate it directly into the model. \citeauthor{li_one_2021} [\citeyear{li_one_2021}] proposed a complex architecture that predicts stock movements by taking as input prices, textual data from the Web, historical data on terrorist incidents, and socio-economic data. Within the architecture, they attempted, from the given inputs, to determine if there has been a significant incident that could impact the market, such as a terrorist attack, and subsequently predict stock movements. The model is interpretable because the final decision is made based on a decision tree. Therefore, users can see the choices the model made to arrive at the prediction. For instance, the interpretations can be : " On 1/6/2009 and 1/7/2009, three news titles are detected to be about terrorist attacks, with key facts including countries (Israel), regions (Gaza), deaths (10, 42, 4), injuries (0, 0, 0), and weapons (firearms, others, others). They are cross-referenced with market and socioeconomic datasets and aggregated into a vector for 1/7/2009. The market is predicted to drop because: 1. S\&P500 index drops today; 2. though the number of deaths is no more than 4, there are attacks in the last two days." \cite{li_one_2021}

Similarly, \citeauthor{carta_explainable_2021} [\citeyear{carta_explainable_2021}] proposed an interpretable approach that includes a decision tree that predicts the magnitude of stock price variations. The approach is divided into four parts: lexicon generation, feature extraction, algorithm prediction and model explanation. The first part takes stock prices and news as input. The correlation between the words found in the news and the movement of prices is computed. This is made by taking the most impactful words into a lexicon for each industry. 
In the second part, the features characterizing the news are extracted. In the third part, the extracted features are fed into the decision tree, which classifies whether the next price variation will be large or small. The final part is to explain the model. It does so in two ways: the rules that determined the predictions are extracted, which is the usual interpretation of decision trees, and lists of the sentence that contains impactful words are extracted.
For instance, sentences containing a lexicon word associated with a significant price variation can be displayed. This offers an additional explanation and a better understanding of the context. These sentences provide clues about whether the result is false or not. The architecture is therefore interpretable because of its feature extraction, the decision tree, and the subsequent interpretation.

\subsubsection{Rule-based}

 Many complex models remain interpretable thanks to rules they contain. These rules often follow the Mamdani-type approach, which involves linguistic variables and fuzzy logic to represent knowledge in a human-readable format. Mamdani-type rules are particularly advantageous for interpretability as they provide  understandable, transparent decision-making processes that can be easily translated into English and understood by humans. Some rules are initiated by experts \cite{yin_interpretable_2023} and others are learned from data with an algorithm \cite{rajab_interpretable_2019, hassanniakalager_conditional_2020, xie_embedded_2022, mingxiang_guo_deep_2021, xie_interpretable_2021, xing_discovering_2019, ferdaus_multiobjective_2022, attanasio_leveraging_2020, cao_multiobjective_2020}.

An interesting rule-based model was introduced in \cite{rajab_interpretable_2019} that proposed  an interpretable neuro-fuzzy approach using Mamdani-type rules for stock price prediction. The model contains a subtractive clustering algorithm \cite{chiu_fuzzy_1994} to craft the rules. The centroids of the clusters become the initial rules. The model's training is done via backpropagation with certain conditions to maintain the model's interpretability. Moreover, a rule selection step is conducted to reduce the number of rules and thus enhance the model's interpretability. The model obtained results comparable to ANFIS \cite{jang_anfis_1993} for the prediction of indices. It strikes a balance between accuracy and interpretability \cite{rajab_interpretable_2019}.

Another way to make an interpretable model based on rules is to integrate rules into an already known system. For instance, fuzzy logic was added to a Hammerstein-Wiener system to form 
the Neural Fuzzy Hammerstein-Wiener (NFHW) network \cite{xie_interpretable_2021}.  It creates Mamdani-type fuzzy decision rules, making it more transparent than a conventional Hammerstein-Wiener. The most frequently used rules are also identified. The model was applied to stock prediction and the results were compared with other regression models based on fuzzy logic. The NFHW outperformed all these models. In another paper \cite{xie_embedded_2022} of the same authors, an interpretable embedded convolutional fuzzy neural network was made. It has a similar architecture than the NFHW, but it now contains a convolutional neural network and a fuzzy association block. The model was tested on financial and biological data. With the financial data, the model predicted the S\&P 500 with better performance than other baseline models in the paper. An example of a Mamdani-type rule given by the model is "If the price on Day 1 is normal, that of Day 2 is low, that of Day 3 is low, that of Day 4 is low, and that of the Current Day is normal, then the closing price of the next trading day is predicted as low" \cite{xie_embedded_2022}. These linguistic rules allow the user to directly understand the decision process behind the predictions.

Instead of relying on a complex architecture, \citeauthor{ferdaus_multiobjective_2022} (\citeyear{ferdaus_multiobjective_2022}) proposed a new algorithm for learning rules, namely the Type-2 Parsimonious Learning Machine (T2-PALM). T2-PALM is an online rule-based neuro-fuzzy learning algorithm designed for predicting fluctuating stock indices. To constrain the number of rules, two multiobjective evolutionary algorithms (MEAs) were employed. The MEAs include a real-coded genetic algorithm and a self-adaptive differential evolution algorithm. The use of these algorithms rendered the model interpretable, while did not deteriorate its accuracy. The PALM is divided into three parts: fuzzyfication, rule addition, and rule merging. Here is an example of rules drawn from \cite{ferdaus_multiobjective_2022}:
\begin{align*}
R_1: \ 
&\text{IF } X \text{ is close to } [0.0006, 0.0723] + [0.1928, 0.9863]x_1 +  [-0.0112, 0.1122]x_2 + \\&[0.0032, 0.0740]x_3 + [-0.0121, 0.1592]x_4 + [0.0329, 0.2562]x_5 \\
&\text{THEN } y_1 = [0.0006, 0.0723] + [0.1928, 0.9863]x_1 + [-0.0112, 0.1122]x_2 + \\&[0.0032, 0.0740]x_3 + [-0.0121, 0.1592]x_4 + [0.0329, 0.2562]x_5
\end{align*}

\begin{figure}[ht]
\begin{center}
    \includegraphics[width=10cm,height=10cm,keepaspectratio]{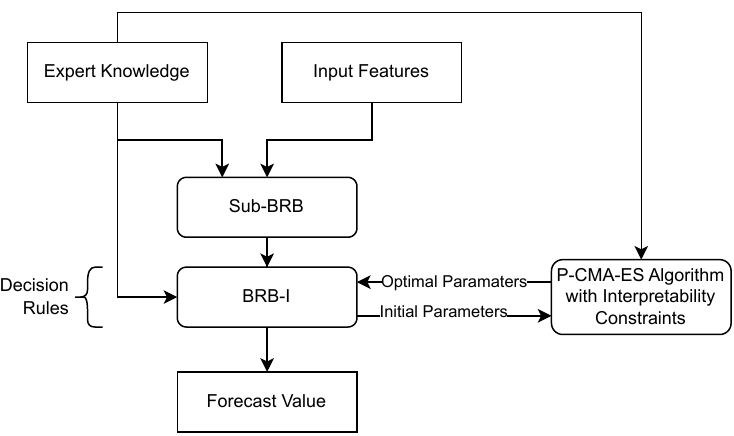}
    \caption{Flow Chart of the Model Presented in \cite{yin_interpretable_2023} }
    \label{fig:HBRB}
    \Description{The Sub-BRB is fed with input features and expert knowledge. The output of the Sub-BRB module is then provided to the BRB-I module. Simultaneously, the BRB module takes expert knowledge as input and receives optimal parameters from the P-CMA-ES Algorithm with Interpretability Constraints module. This algorithm, taking expert knowledge and initial parameters from BRB-I as input, outputs the optimal parameters. The final forecast value is obtained from the BRB-I module.}
\end{center}
\end{figure}

In contrast of the other models, the rules of Hierarchical Belief Rule Base (HBRB-I)  \cite{yin_interpretable_2023} come initially from experts. Indeed, HBRB-I is composed of multiple Belief Rule Base (BRB) modules, a decision-making system based on rules. The HBRB is used to predict stock movements. There are 11 specific criteria related to modelling, inference, and optimization to enhance the interpretability of the model. These criteria suggest that the system should be simple and consistent, and have a clear semantics. They also suggest, among other things, that the inference process should be transparent and "the expert knowledge should be reasonably used "\cite{yin_interpretable_2023}. All these criteria are written in a mathematical form. Within the HBRB, the different rules of the BRBs are initiated with the help of experts. Subsequently, the BRB-I module learns to manage these rules by learning the right weights to properly weigh them. The weights of these rules can also be analyzed according to their importance. The architecture of the HBRB-1 is presented in the Figure~\ref{fig:HBRB}.

Finally, fuzzy logic can be combined with (rough) neurons into a neural network. It is the case of the Fuzzy Rough Neural Network (FRNN) from \cite{Shihabudheen_recent_2018} and \cite{gu_particle_2021}.  \citeauthor{cao_multiobjective_2020} [\citeyear{cao_multiobjective_2020}] modified the FRNN to enhance its performance to predict stock prices. The model is inherently interpretable due to the fuzzy logic it is built upon. A layer of the network consists of IF-THEN rules. The rules used for decision-making can thus be extracted from the network, providing insight into the computation process.

\subsection{Trends}
\label{TS_interpretation}

Model interpretability has also been investigated from the perspective of trend analysis. Indeed, the trend is an important component of a time series that can be decomposed and analyzed in depth to extract significant information. The models presented in this section analyze time series to identify trends in the data, characterize them, and then link them to the prediction. 

A GRU method based on a HPFilter was introduced in \cite{xiong_gru_2021} to predict stock prices. The HPFilter is an approach employed in macroeconomics for trend analysis. In the architecture, it allows for the decomposition of opening prices, closing prices, minimum prices and maximum prices into two components, i.e. long-term trend and short-term volatility. Each of these components is modelled by a specific GRU: SGRU for short-term and LGRU for long-term. The output of the components are combined to form the final prediction. 
 This architecture merges the concept that a time series is composed of trends and volatility with the high performance of GRUs, achieving both accuracy and interpretability. Indeed, the model is more transparent than deep neural networks due to the trend decomposition, which allows the short-term fluctuations and the long-term trend to be visualized. Also, the combination of two models, with one GRU predicting the trend and another predicting the volatility, surpasses the accuracy of a single GRU \cite{xiong_gru_2021}.

Similarly, the N-BEATS model proposed in \cite{oreshkin_n-beats_2020} allows the interpretations of trends and the seasonality.
In general, N-BEATS, designed for univariate time series prediction, is not interpretable due to its complex architecture. However, the authors proposed a special configuration with constraint aligned with trends and the seasonality of the time series, making the model interpretable without decreasing the results. Indeed, this interpretable configuration allows a trend and seasonality decomposition. The seasonality and trends were separately extracted from the model's predictions, and then analysed in charts alongside with the predictions.

Just as trends compose time series, linear and nonlinear relationships also form time series. This concept is known as "hybrid prediction" \cite{zhang_time_2003}. According to this approach, 
a time series can be represented as the sum of a linear component and a non-linear one. The linear component and the non-linear component can be extracted from the time series to provide interpretation for the prediction. The linear part is modelled with ARIMA, while the nonlinear part is modelled with a neural network.
Residuals are defined as the difference between the actual prediction and the linear part. \citeauthor{panja_parnn_2022} (\citeyear{panja_parnn_2022}) adopted this approach in their own way. They introduced the Probabilistic Autoregressive Neural Network Framework (PARNN) for time series prediction. This model computes the linear part of the input data using ARIMA and then calculates the residuals. Subsequently, the residuals and the input time series are fed into a neural network with one hidden layer and one output neuron. The output of this neural network is the time series prediction.
 This model is interpretable due to the transparent use of the two modules that compose it. As ARIMA is interpretable based on its functioning, so is this model by transition. Indeed, ARIMA provides information on autoregression, moving average, and integration.  We recognize that this model is not the most interpretable one. However, since ARIMA provides some insights on the inputs, we classify it as interpretable, even though predicting a time series with ARIMA alongside any other model would give similar explanations.

\subsection{Interpretability Evaluation}
\label{inter_eval}
How to evaluate interpretability is an open issue in the machine learning community. Existing approaches to evaluating the interpretability deal mainly with one of the two following aspects: the interpretability of a model, i.e. how much the model is interpretable, and the reliability of the interpretations, i.e. how closely the XAI principles reflect the underlying logic of the model.

Ideally, we would like to have some metrics that allow to directly measure the level of interpretability of a model. However, according to our knowledge, there is no such metrics function capable of doing that. Instead, existing work such as that in \cite{lipton_mythos_2018} suggests three interpretability dimensions to be considered.
The first one is the simulatability. This concept refers to the ability of a model to be entirely inspected by a human at once. It is the simplicity of a model. For example, a linear model with low-dimensional features is more interpretable than a linear model with high-dimensional features according to simulatability. This concept can be a bit subjective, because everyone does not have the same level of analysis capacity. Indeed, if we want to measure the simulatability, we need to quantify the properties (number of parameters, number of sub-modules, number of layers, number of different sub-model, etc.) of the model. Almost all the models presented in this survey would have a low score in simulatability due to their complexity and their high number of modules. 
The second dimension proposed by \cite{lipton_mythos_2018} is the decomposability. It is the ability of the parts (input, parameters and calculations) of a model having an intuitive explanation. For example, the parameters of the linear models represent the strength (feature importance) of the association between the features and the prediction. Another example is the nodes of decision trees that represent interpretable rules.
The decomposability requires the input to be intuitively interpretable. Indeed, it would be hard to analyze the importance of a feature that is non-interpretable itself. 
The interpretable models presented in this paper would have different scores of decomposability, because the percentage of interpretable parts is different among different models. 
The third dimension of interpretability according to \cite{lipton_mythos_2018} is the algorithm transparency. This notion is the ability of the learning algorithm of a model to be simple, understandable by humans. Also, it includes the ability of a model to have a unique solution. As an example, linear models have this ability, unlike deep neural networks.  

Evaluation of the reliability of the information provided by interpretable models is also a challenge. Since the models employed in practice are usually complex,  the interpretations from such models may not be reliable. In fact, as we discussed earlier, in a complex interpretable model, the XAI principles result from the use of some technique that is placed in some part of the model's decision process. When this part is directly linked to prediction, such as at the end of the decision process as presented in the \cite{munkhdalai_recurrent_2022}, it allows illustrating how the prediction is made. For example, if the prediction is computed by a linear sum that includes interpretable features, either internal or from input, the weights of the linear sum directly represent the importance of these features.  
While the process to obtain the feature weights may be incomprehensible (opaque), the relationship between the features and the prediction can be understood without ambiguity.
As long as the features mean something for the user, these weights appear to be reliable.

However, if the use of the technique is followed by several other operations, the reliability of the information generated from the technique can be seriously affected. Indeed, sometimes the  component that measures feature importance is at the beginning of the model's process, for example, the attention mechanism as in \cite{zhang_at-lstm_2019}. This component allows the model to weight the input before processing it and makes the model interpretable As the authors in \cite{lin_kernel-based_2022, cheng_financial_2022, zhou_domain_2020, guo_exploring_2019, zhang_at-lstm_2019, yang_explainable_2019, chang_memory-network_2018, tran_temporal_2019} suggested, these weights could be extracted and analyzed as the feature importance. Nevertheless, since processing steps are operated between this interpretable component and the final prediction, it is questioned about whether the feature importance will remain consistent after all these operations.  It is possible that the model might learn to ignore those weights and instead learn the patterns of the data in another way. Consequently, the feature importance might have changed after these operations. This means that even if there's a module responsible for weighting the input at the beginning of the model, subsequent operations can change this weighting when it comes to prediction. Therefore, even though neural networks having an attention mechanism at the beginning like \citeauthor{zhang_at-lstm_2019} [\citeyear{zhang_at-lstm_2019}] are considered interpretable, the reliability of the interpretation should be questioned.

Going somewhat beyond the time series models, several interesting works have been done by the NLP community on how to determine if the attention weights truly represent the feature importance. Their conclusions are not affirmative. Indeed, the attention weights are not considered as the best option for ranking features according to \cite{serrano_is_2019}. The authors of this paper show that the attention weights contain some useful information about the feature importance, but the information may not be enough to correctly rank the features. Also, the capacity of the attention weights is also questioned in
\cite{jain_attention_2019}, where the attention weights do not confirm the feature importance.  Indeed, feature importance from attention weights shows a low correlation with the one obtained from other methods, such as gradient-based and leave-one-out methods.
Even if benchmark methods do not represent the truth,  the experience shows that the reliability of the attention weights is questionable. If we extend the logic of the definition of the "explanation" in \cite{wiegreffe_attention_2019}, the attention weights could be reliable, depending on the definition of the reliability. If a reliable interpretation has to be plausible and faithful, attention weights would not be reliable. However, if it only has to be plausible, they would be one. These divergent results 
suggest that the reliability of the attention weights depends on the context, the definition and the testing method. It could thus be different in a task of forecasting time series.  These analyses show that the reliability of the interpretation of complex models must be challenged, and their use must be done with caution.

In general, there is no standard method to measure the reliability of the interpretations. However, at this point, the interpretations of an interpretable model and the explanations of an explainable model are essentially equivalent, because the difference between an interpretation and an explanation results from their source, not from their nature. Some ideas and approaches to evaluate explanations are presented in Section \ref{explain_eval}. They could be applied to interpretations.

\section{Explainable Models and Explainability Methods}
\label{explainable}

In this section, explainability methods and their context of application, including the models to be explained, are presented. Since this paper focuses on XAI, much emphasis has been placed on the explainability methods rather than on the models themselves. Some financial insights that the authors derived from these methods are also presented. The explainability methods are grouped according to their XAI principles, such as feature importance and visual explanation, and according to the technique that they use to extract explanations from models. Finally, issues relating to evaluation of explainability are discussed.

\subsection{Feature and Time Importance}
With a similar objective as interpretable models, explainable models employ a technique to elucidate the impact of features on the output. Basically, two techniques are used to measure the feature importance of models: the propagation technique and the perturbation technique. Methods using the propagation technique track features through the model to gather information about their influence and observe the outcomes, while methods using the perturbation technique alter the inputs and compute the difference in results in order to quantify the effect of the features.

\subsubsection{Propagation}

The propagation technique, designed primarily for analyzing neural network based models, makes use of the information provided by the gradient or the neurons. 
Indeed, neural networks fundamentally rely on a sequence of weighted multiplications. These multiplications can be followed through the model from the input layer to the output layer during the forward pass. They can also be traced from the output to the input layer during the backward pass. The composition of multiplication allows neural networks to be differentiable. The propagation technique is based on these properties, i.e. the multiplicative composition and the differentiability of neural networks. Similar to the propagation of information in the forward and backward passes of neural networks, the XAI methods using propagation can compute a score for each neuron from the output layer to the features layer to determine the importance of the features. An example of such a propagation technique is Layer-Wise Relevance Propagation (LRP) proposed by \citeauthor{bach_pixel-wise_2015}\cite{bach_pixel-wise_2015}. Moreover, these XAI methods can also take advantage of these passes to compute feature importance such as Integrated Gradient \cite{sundararajan_axiomatic_2017} thanks to the differentiability of neural networks.

A neural network for predicting stock returns, named Deep Factor network, was proposed in \cite{nakagawa_deep_2019} that employed LRP \cite{bach_pixel-wise_2015} as the explainability method.  
LRP propagates relevance scores, representing the feature importance, from the output layer of the network back to the input layer like the backpropagation. The relevance scores of neurons of each layer are computed from the weights of the layer and the relevance scores of the previous layer \cite{nakagawa_deep_2019, bach_pixel-wise_2015}. The final result of this propagation is the feature relevance/importance of financial factors including risk, quality, momentum, value, and size as the input features were grouped by financial themes (factors). 
The feature importance given by the LRP has been compared with correlation coefficients including Kendall and Spearman.
Their rankings did not align. For example, LRP indicated a high contribution from the value factor, while the correlation coefficients suggested low impact. 
This verification does not affirm or dismiss the reliability of LRP, but it highlights a divergence between the two methods. As an explainability method, LRP will need more rigorous testing in a financial context.
The Deep Factor network has also been compared to other models, including linear ones and proved to be more accurate. Finally, the work of \cite{nakagawa_deep_2019} include also a method to determine the importance of the financial factors through different time periods.

 As mentioned earlier, integrated gradient \cite{sundararajan_axiomatic_2017} is another explanation technique that uses the differentiability of the neural networks to explain them. Indeed, in its process, the partial derivatives of the model are computed. The method consists in computing the integral of the partial derivative along the line between the input (feature) point $x$ and a baseline point $x'$, which, for time series prediction, is defined as the average of $n$ most recent data points. 
The integral computes the cumulative contribution of a feature from the baseline point to the input point. Since the sum of the integrated gradients of all features equals the difference between the prediction at the input point and the prediction at the baseline point ($ \sum_i^n IG_i(x) = F(x) -F(x')$) \cite{sundararajan_axiomatic_2017}, the integrated gradient of a specific feature can be interpreted as the component that contributes to adjusting the model's prediction from $x'$ to 
$x$. The results of the method then depends on this baseline point \cite{freeborough_investigating_2022}. Therefore, with $\frac{\partial F(x)}{\partial x_i}$ being the gradient component of the neural network $F$ along the $i^{th}$ dimension, the integrated gradients measuring the importance of the feature $i$ is defined by \cite{sundararajan_axiomatic_2017}:

\begin{equation}
    IG_i(x) ::= (x_i - x_i') \times \int_{\alpha = 0}^{1} \frac{\partial F (x' + \alpha \times (x - x'))}{\partial x_i} d\alpha
\end{equation}

\noindent Integrated gradient has been applied to explain the hierarchical neural network in \cite{chiewhawan_explainable_2020}. This hierarchical neural network takes both textual data and numerical data as input to predict the SET (Thailand stock market) index, which causes difficulties to LRP based method for model explanation. The integrated gradient allows to extract the input words that positively and negatively impacted the stock prediction.

\subsubsection{Perturbation technique}

The perturbation is a well-known technique for obtaining the feature importance of a black box model. The concept behind this technique is straightforward. Firstly, a prediction is made using all features of the trained model for a given input point. Then, within the set of all features, the feature of interest is perturbed, and another prediction is generated using this perturbed feature along with all other features for the same input point. A comparison, basically in terms of their differences, between these two predictions is made to estimate the importance of the features. The methods that employ the perturbation technique can be classified into classic methods and sophisticated methods. The classic methods are the ones that apply directly this concept and the sophisticated methods rely on the perturbation technique. The sophisticated methods include SHAP \cite{freeborough_investigating_2022} and LIME \cite{ribeiro_why_2016}.

In general, the methods using perturbation are easy to apply. Indeed, they are model agnostic, i.e. they do not depend on the model, so they can be applied on all types of models. The classic methods are easy to code, while the sophisticated ones, like SHAP or LIME, have their library already implemented in Python, which makes them readily applicable. The output of these methods is the feature importance that can help understand the relation between the features and the output of a model. It can also be used to make feature selection according to \cite{carta_explainable_2022}.  As an example, the popular and highly effective model for predicting financial time series, XGboost, has been explained by SHAP \cite{weng_analysis_2022, ghatnekar_explainable_2021, deng_stock_2023}. The perturbation technique can also be used to compute a confidence level of a prediction \cite{celik_extending_2023}.

\paragraph{Classic Pertubation Methods} As we mentioned earlier, the idea behind the classic perturbation methods is to modify a feature, make a prediction, and then compare the result with the prediction using the original feature through a dissimilarity measure. This process can be applied to a single prediction or a set of predictions. There are several ways to modify the feature. The first approach involves randomly shuffling the data of the feature with historical feature values, known as Permutation Importance, as proposed by \cite{breiman_random_2001}. The second method is to add noise to the feature values, as discussed by \cite{carta_explainable_2022}. Determining the appropriate magnitude of the noise can be challenging for this method. The third method is to zero out the feature, referred to as "ablation" \cite{freeborough_investigating_2022}. In the context of time series, setting a feature to zero may not be meaningful. The common solution to this issue is to replace the value with the average value of the feature over the last $n$ days. 

Classic perturbation methods, such as ablation, added noise, and permutation importance, were applied on recurrent networks (RNN, LSTM, and GRU) in  \cite{freeborough_investigating_2022} and in \cite{cascarino_explainable_2022}. We discuss here the work reported in \cite{freeborough_investigating_2022}. The models take, as input, high, open, low close, volume and adjusted close to predict stock prices. The integrated gradient method was also tested. The baseline used for IG was the average of the last 91 days, on the gradients. 
Based on the findings, the authors concluded that these techniques are applicable to recurrent networks employed in financial time series prediction to provide explanation of their complex architecture.
Also, they found that the methods are complimentary. It makes sense because each method was applied in a way to extract some different information. Indeed, the ablation and IG aim to extract the time and feature importance. More specifically, it was possible to determine with IG the sign of the correlation between the feature at a specific time with the output, but not with the ablation method. The added noise was used to extract only the time importance, and the permutation importance only extracted the feature importance. The results of all the four methods show that the importance of the volume is low as compared to the other features. The importance of the open, high, low, close and adjusted closed features are mostly equal for all models and methods. This is explained by the fact that the volume does not refer to the price, but to the number of stocks traded. 
Indeed, if a stock is bought or sold, the volume value will increase, giving no information about the price of the stock. Also, as expected, the most recent days closest to the prediction are more important than the more distant ones according to ablation, integrated gradients and added noise. The authors have presented a complete analysis of these methods through the different models. The importances of ablation and permutation importance remain relatively constant across the models, while the effects of IG and added noised vary significantly. 

\paragraph{SHAP} SHAP (SHapley Additive exPlanations) is a sophisticated method for explaining the output of machine learning models and is based on the perturbation \cite{freeborough_investigating_2022}. It is rooted in game theory, where it assigns each feature an importance value for a particular prediction. SHAP evaluates the effect of a feature by considering its absence in all possible combinations of features and calculates the difference in predictions compared to when the feature is present. The values computed by SHAP, named SHAP values, measure the average contribution of a feature to every prediction, ensuring consistent and fairly distributed feature importance scores. These values can be computed for any model and provide intuitive insights into its behaviour. By decomposing a prediction into the sum of SHAP values for each feature, one can understand how differently features contribute, either positively or negatively, to the prediction. It was proved that SHAP is the only additive model that respects the properties of local accuracy, missingness and consistency \cite{lundberg_local_2020}. 

There are several implemented versions of SHAP depending on the model to which it is applied. For example, Kernel SHAP can be applied on all types of models. However, the independence between features is assumed to compute Kernel SHAP. Another example is TreeSHAP \cite{lundberg_local_2020}, which is the SHAP method adapted for tree-based models. As the tree gradient boosting model XGboost is popular in financial time series forecasting, TreeSHAP is useful to compute its feature importance. Indeed, \citeauthor{weng_analysis_2022} [\citeyear{weng_analysis_2022}] analyzed the role of financial pressure in predicting the volatility of healthcare stocks with tree-based models, including XGBoost, and TreeSHAP. The models take as input past volatility of health care stocks from China and USA and the financial pressure measured by Office of Financial Research Financial Pressure Indexes (OFR FSI). The OFR FSI contains five financial categories such as credit, equity valuations, safe assets, funding deviation and volatility. The target value was the future volatility of health care stock of both countries. TreeSHAP was used to calculate the relative feature effect, that is, the SHAP values divided by a benchmark. The relative feature effect was utilized in four analyses. 

Firstly, the effect of financial pressure (OFR FSI) on the volatility of health care stocks was analyzed for China and USA. The health care stocks of China seemed to be more sensitive to financial pressure than the health care stocks of the USA due to their greater contribution to the prediction. 
Based on the theory of \cite{floro_threshold_2017,macdonald_volatility_2018}, the authors explained this result by the fact that "emerging countries pursue quicker and aggressive contingent response strategies in response to external financial pressures." \cite{weng_analysis_2022}. 
Secondly, the importance of the five categories of financial pressure on the volatility of health care stocks was computed for the USA and China. The equity indicator was found the most important indicator for both countries, while the safe asset indicator is the least important one. 
The authors explained that by the fact that "the extent of investor confidence and risk appetite directly results in the extent of market illiquidity and increased asymmetric information"\cite{weng_analysis_2022}, according to \cite{craig_financial_2009} , and also that the higher the cash flow, the more stable the stock market will be. 
Thirdly, the different period lags of the OFR FSI were analyzed. The 7-period lag was the most important for both countries, but the second most important was 6-period lag for the China and 1-period lag for the USA. It shows that countries respond at different speeds to financial pressure \cite{weng_analysis_2022}. Fourthly, the contribution of the 7-period lag of the five financial pressure indicators was calculated. It shows that the distribution of the contribution of different period lags changes through indicators. Indeed, the effect of the different period lags is similar to the credit, while it is very heterogeneous for the equity indicators for both countries. However, the period lag for each indicator is different for the USA compared to China. Indeed, they found that "the volatility forecasting of Chinese health care stocks is mainly influenced by each indicator of financial pressure at the late and medium period lags; however, it was influenced at the front and medium period lags for the USA." \cite{weng_analysis_2022} 

These analyses are concrete examples of how we can use XAI to understand financial patterns. Originally, SHAP computes local feature importance. Then, global feature importance can be computed by doing the average of the SHAP values over the training sample of a model like\cite{xie_forecasting_2022} did to understand which features were important in the training of their model and to do a feature selection after. Other examples of use of SHAP to explain a model for predicting financial time series are presented in \cite{ghosh_hybrid_2022, yang_interpretable_2023, jaeger_interpretable_2021, ohana_explainable_2021, ghosh_prediction_2023, deng_stock_2023, gradojevic_unlocking_2022, neghab_explaining_2023, ghatnekar_explainable_2021}.

\paragraph{LIME} LIME (Local Interpretable Model-agnostic Explanations) is a method designed by \cite{ribeiro_why_2016} to explain the predictions of any black-box machine learning model. It works by approximating the complex model locally around the prediction of interest using an interpretable model, typically a linear model. It supposes that the surrogate and the black-box models are equal at the point of interest. By perturbing the input data, generating new samples with perturbation and the black-box model, fitting a surrogate model and interpreting the surrogate model, the black-box model can be explained. This provides a human-understandable explanation, showing how each feature contributes to the model's decision for a given instance. 

Contrary to SHAP, LIME cannot be used to explain a model globally. This is one of the reasons why \citeauthor{ghosh_covid-19_2023} [\citeyear{ghosh_covid-19_2023}] used LIME to obtains local feature importance and SHAP to offer the global feature importance. Their goal was to predict Spanish and Indian stock during the COVID-19 pandemic and to understand the effects of macroeconomic indicators and media chatter on these predictions. The media chatter-linked is measured by some indices such as Panic Index and Media Hype Index. Many financial conclusions were made by the authors. Indeed, they found that the Spanish stocks seem to be less predictable than the Indians stocks. There would be a higher inefficiency in the Indian market than in the Spanish one. Also, they found that the global and local macroeconomic indicators were important in the prediction. The authors suggested that the media chatter should also be considered to "decode the temporal evolutionary pattern of the underlying assets." \cite{ghosh_covid-19_2023}. Even though the pandemic brought about significant disruptions and uncertainties, it did not fundamentally alter the underlying inefficiencies that exist in financial markets due to the importance of macroeconomics reflectors, but the media chatter emerges as an important feature in the prediction of the stocks of the two countries. Indeed, media hype and sentiment should be considered when making short-term investments in the equity market \cite{ghosh_covid-19_2023}. This work highlights the benefits of using LIME to understand  financial contexts.

An issue with LIME is that it is hard to define the region around the prediction, especially when it is a time series. That is why \cite{schlegel_ts-mule_2021} introduced a new technique, named TS-MULE for segmenting and replacing the time series. Another version of LIME was presented in \cite{celik_extending_2023} to apply it in an unconventional manner. Indeed, it is used not to show feature importance, but to transform the binary output of the random forest into an "explanation prediction probability". 
This probability allows assigning a confidence level to the base classification. It informs the model and the user whether to rely on the prediction based on a chosen confidence threshold. As the confidence level is used to make final predictions, the architecture could be classified as interpretable. However, it would make more sense to put it in the LIME section with the other variants of the LIME method. Furthermore, another special case of LIME is when a decision tree is introduced in the surrogate model. In this case, in addition to generating feature importance, this model has the ability to produce local decision rules \cite{lahiri_accurate_2020, qi_short-term_2020}. For instance, in \cite{qi_short-term_2020}, the rules were extracted and tested according to their investment utility. The best rules were kept according to this test.

\subsection{Visual Explanations}

Visualizing the inputs, core, or outputs of a model is an effective way to understand the model and its predictions. This principle, named visual explanations, regroups all visual information that helps the user comprehend the model. In general, this principle is inherently understandable to humans due to their visual nature. Compared to other XAI principles that focus on the information presented, this one focuses on how the information is presented to be human understandable. It is often accompanied by other principles, like the feature importance. The technique presented in this survey for providing the visual explanations is the user interface (UI). In this context, the UI is defined as an application showing explanation of a model. The UI often contains multiple types of information, like graphs, plots, words, table, etc. It can show information about the model itself, such as parameters or feature importance, but it can also contain complementary information that helps the user understand the model, the prediction and the context. This information includes plots of past data, news updates, comparison with past prediction or real data, model details, etc. It regroups all kinds of information that helps the user in its work of data scientists. Another technique of visual explanations would be the dimension reduction, as applied in \cite{kim_stock_2018}. The concept aims to reduce the dimensionality of the multivariate time series to be able to visualize it in two or three dimensions.

User interfaces are useful for financial data scientists. They aim to provide a clear vision of the model and other contextual aspects. An interface can offer a clear aspect of the financial context. Indeed, it can present the stock's historical data to observe the events leading up to the last period of the prediction, thereby allowing users to evaluate the validity of the prediction. Additionally, technical indicators could be computed and displayed on the interface for further financial analysis. To visualize the outputs of a prediction, it's useful to plot them against real values for accuracy evaluation or against benchmarks for profitability analysis and display them on the interface. Considering the predictive power of the news, some architectures use them as input. A user interface can be used to show news and to show related ones to better understand the financial context. The UI can also be made interactive to enhance the user experience by allowing customization of model visualization and parameter settings. The details of a model, such as parameters, loss, and error, are useful for the user to understand the prediction and the model itself. Ultimately, the user interface is a highly flexible approach to understand the model and the context in order to make more enlightened decisions in a financial environment. 

DeepVIX \cite{dang_deepvix_2020} is an interactive interface designed to explain LSTM networks. The interface shows five different information: training parameters, comparison between training and test loss, configuration parameters, LSTM architecture, and interactive features. In fact, the interactive features part allows users to view the four types of weights: input gate, forget gate, cell state, and output gate, throughout the network and iterations. It literally shows the weights of the LSTM through a plot to correspond to the shape of the network. Its primary aim is to offer a deeper understanding of the LSTM, enabling better optimization. With this interface, it is easy to observe changes in the parameters throughout the iterations and relate these changes to machine learning theory.  

Another popular interactive interface  is DeepClue, as introduced in \cite{shi_deepclue_2019}. It's a visual interface designed to elucidate the hierarchical neural network also proposed in  that paper. This network takes financial news headlines as input and predicts stock prices. Using the LRP \cite{bach_pixel-wise_2015}, a relevance score is computed to measure the feature importance. 
The interface allows users to visualize these relevance scores, as well as to view stock timelines, document lists, keywords map, and more. It also displays a list of documents related to the selected factors, providing a comprehensible view of the context to increase understanding of the model. Since the inputs are text, the explanations become inherently clearer for the user because the inputs are interpretable themselves. DeepClue helps to link the model's decisions to financial theory.  It was used to upgrade the model and to verify a prediction. Indeed, with the help of a trader and a stock analyst, the model's decisions were evaluated in \cite{shi_deepclue_2019}.  The trader analyzed the best moments to make buy/sell a specific stock , namely during significant increases or decreases.  
He analyzed the important words and identified a key cluster of words that made sense for him and for the prediction. The interface showed him the news from the cluster's source, thus enabling verification of the prediction's validity.  The analyst, on the other hand, found a deficiency in the model with the help of DeepClue: "the afterhour trading is not considered in the training phase" \cite{shi_deepclue_2019}. Furthermore, Deepclue also helped the analyst discover that the model was overfitting in a case where the inputs were social media words. It's a true example where XAI increases the user's confidence in the model by linking the prediction to financial theory. 
Two other interfaces are Hawkeye from \cite{carta_hawkeye_2021} and the one presented in \cite{bandi_integrated_2021}.

\subsection{Explainability Evaluation}
\label{explain_eval}
 Similar to interpretability evaluation, there is no established standard for explainability evaluation. As the concept of evaluation is large, it must be defined. Also, the evaluation could depend on the XAI principles of the methods. In this subsection, we present a few main criteria and approaches in this area.

According to \cite{belle_principles_2021}, we should consider evaluating the explainability on the following criteria: comprehensibility, fidelity, accuracy, scalability and generality. These criteria were defined initially in the context of rule extraction by \cite{shavlik_rule_1999}. They are theoretical concepts that make sense,  but are difficult to quantify. We could add to these criteria the ability of an explanation to be adapted to the context and to the user. Another way to evaluate the explainability is to inspect the effect of the explainable method on a real task. Indeed, a study by \cite{cau_supporting_2023} on the trust in XAI during high-uncertainty decisions was made. The authors examined the effects of AI explanations on user trust when the AI displays high versus low confidence levels. The study compared three types of explanations: inductive (based on local examples), abductive (highlighting local feature importance), and deductive (local rule-based explanations). Their findings suggest that explanations diminish performance when the model projects low confidence. Conversely, in scenarios where the model is highly confident, both abductive and deductive explanations enhance the user task performance compared to inductive explanations.

Two interesting measures for evaluating explainability techniques that show the importance of features in a time series prediction context are introduced in  \cite{ozyegen_evaluation_2022}. They are named "Area Over the Perturbation Curve for Regression (AOPCR)" and "Ablation Percentage Threshold (APT)".
The AOPCR measures the effect of removing a certain number of features, while the APT measures the percentage of features that need to be removed before crossing a certain threshold.
Using these measures, the authors compared SHAP with intuitive methods, namely a local average replacement explainability method, a global average replacement explainability method, and a random method.
Specifically, the values of importance of features calculated by these methods were compared. The analysis was performed on three different models: a Time Delay Neural Network, an LSTM, and a Gradient Boosted Regressor.
They concluded that, in general, SHAP outperformed the other methods according to their measure.

\section{Further Characterization of XAI Models}
\label{alt_taxonomy}

\citeauthor{das_opportunities_2020} [\citeyear{das_opportunities_2020}], \citeauthor{kumar_overview_2023} [\citeyear{kumar_overview_2023}] and  \citeauthor{banerjee_methods_2023}[\citeyear{banerjee_methods_2023}] classified explainability techniques based on three criteria: methodology, usage, and scope. Here, the methodology refers to the interpretable or explainable technique, the usage describes how the method is applied, whether post-hoc (\textbf{PH}) or intrinsic (\textbf{IN}), while the scope of the explanation can either be global (\textbf{GB}) or local (\textbf{LO}). All the interpretable models are intrinsic and all explainability methods are post-hoc. The usage criterion is unconsidered 
for this reason. A local explanation is an explanation for one prediction, and a global explanation is an explanation for an entire dataset. Contrary to \cite{das_opportunities_2020} that only categorizes  explainability methods, we also categorize interpretable models. Furthermore, the XAI principles defined earlier are included in this charaterization. 
They are the feature importance (\textbf{FI}), the confidence level (\textbf{CL}), decision rules (\textbf{DR}), visual explanation (\textbf{VE}) or trends (\textbf{T}). The Table~\ref{tab:alt_tax} shows explainable methods and interpretable models categorized according to these criteria. The Table~\ref{tab:abbreviation} contains all the abbreviations of the Table \ref{tab:alt_tax}. Additionally, some papers we reviewed are listed in the table even though they are not explicitly described in the survey, as they did not meet the criteria of the methodology.


\begin{table*}[ht]
	\caption{Characterization Table}
	\label{tab:alt_tax}
	\scalebox{0.72}{
		\begin{tabular}{lllllllll}

			\toprule
			XAI Method                 & Pub.                                              & Year                                                                                            & Type & XAI Prin. & Scope  & Meth.     & Model Dep. \\
			\midrule

			LIME                       & \cite{ribeiro_why_2016}                           & 2016  & E    & FI        & LO     & PER, APP     & MA         \\
			TreeSHAP                   & \cite{lundberg_local_2020}                        & 2020  & E    & FI        & LO, GB & PER          & MS         \\
			SHAP                       & \cite{lundberg_unified_2017}                      & 2017  & E    & FI        & LO, GB & PER          & MA         \\
			LIME (non-usual)           & \cite{celik_extending_2023}                       & 2023  & E    & CL        & LO     & PER          & MS         \\
			DEMUX                      & \cite{doddaiah_class-specific_2022}               & 2022  & E    & FI        & LO     & PER          & MA         \\
			Dynamask  & \cite{crabbe_explaining_2021}                     & 2021               & E    & FI        & LO     & PER          & MA         \\
			DeepVix                    & \cite{dang_deepvix_2020}                          & 2020  & E    & FI        & GB     & VE           & MS         \\
			LRP                        & \cite{bach_pixel-wise_2015}                       & 2015  & E    & FI        & LO     & PRO          & MS         \\
			DeepClue                   & \cite{shi_deepclue_2019}                          & 2019  & E    & VE        & GB     & VE           & MS         \\
			Integrated Gradient        & \cite{sundararajan_axiomatic_2017}                & 2017  & E    & FI        & GB     & PRO          & MS         \\
			Permutation Imp.           & \cite{altmann_permutation_2010}                   & 2010  & E    & FI        & LO, GB & PER          & MA         \\
			Ablation, Noising          & ~                                                 & ~     & E    & FI        & LO, GB & PER          & MA         \\
			LMTE                       & \cite{lahiri_accurate_2020}                       & 2020  & E    & FI        & LO     & PER          & MA         \\
			LSTM-Forest                & \cite{park_stock_2022}                            & 2022  & E    & FI        & GB     & PER          & MS         \\
			ALE       & \cite{apley_visualizing_2019}                                      & 2019                       & E    & FI        & LO, GB & PER          & MA         \\
			ALE range & \cite{liang_time-sequencing_2022}                                  & 2022                     & E    & FI        & LO, GB & PER          & MA         \\
			Hawkeye                    & \cite{carta_hawkeye_2021}                         & 2021  & E    & VE        & LO, GB & OTH          & MA         \\
			~                          & \cite{bandi_integrated_2021}                      & 2021  & E    & VE        & LO, GB & OTH          & MA         \\
			GRU-TREE                   & \cite{wu_beyond_2017}, \cite{wu_preliminary_2019} & 2017  & I    & DR        & LO, GB & DT           & MS         \\
			~                          & \cite{munkhdalai_recurrent_2022}                  & 2022  & I    & FI        & LO     & LR           & MS         \\
			-                          & \cite{lin_kernel-based_2022}                      & 2022    & I    & FI        & LO, GB & ATT          & MS         \\
			CF                         & \cite{hassanniakalager_conditional_2020}          & 2020  & I    & DR        & LO, GB & FL, DR       & MS         \\
			~                          & \cite{xie_embedded_2022}                          & 2022  & I    & DR        & LO, GB & FL, DR       & MS         \\
			HBRB-I                     & \cite{yin_interpretable_2023}                     & 2023  & I    & DR        & LO, GB & DR, FL       & MS         \\
			NFHW                       & \cite{xie_interpretable_2021}                     & 2021  & I    & DR        & LO, GB & DR, FL       & MS         \\
			~                          & \cite{rajab_interpretable_2019}                   & 2019  & I    & DR        & LO, GB & DR, FL       & MS         \\
			DCAFS                      & \cite{mingxiang_guo_deep_2021}                    & 2021  & I    & DR        & LO, GB & DR, FL       & MS         \\
			~                          & \cite{xing_discovering_2019}                      & 2019  & I    & DR        & LO, GB & OTH          & MS         \\
			~                          & \cite{carta_explainable_2021}                     & 2021  & I    & DR        & LO, GB & DT           & MS         \\
			N-BEATS                    & \cite{oreshkin_n-beats_2020}                      & 2020  & I    & T       & LO     & OTH          & MS         \\
			PARNN                      & \cite{panja_parnn_2022}                           & 2022  & I    & T       & GB     & OTH          & MS         \\
			~                          & \cite{xiong_gru_2021}                             & 2021  & I    & T       & GB     & OTH          & MS         \\
			~                          & \cite{zheng_finbrain_2019}                        & 2019  & I    & FI        & LO, GB & FIN          & MS         \\
			~                          & \cite{li_incorporating_2020}                      & 2020  & I    & FI        & GB     & GRA          & MS         \\
			KDTCN                      & \cite{deng_knowledge-driven_2019}                 & 2019  & I    & FI        & GB     & GRA          & MS         \\
			~                          & \cite{luo_learning_2022}                          & 2022  & I    & FI        & GB     & LR           & MS         \\
			$L^3$                      & \cite{attanasio_leveraging_2020}                  & 2020  & I    & DR        & GB     & DR           & MS         \\
			ML-GAT                     & \cite{huang_ml-gatmultilevel_2022}                & 2022  & I    & FI       & GB     & GRA          & MS         \\
			T2-PALM                    & \cite{ferdaus_multiobjective_2022}                & 2022  & I    & DR        & LO, GB & DR           & MS         \\
			FRNN                       & \cite{cao_multiobjective_2020}                    & 2020                                                                                                 & I    & DR        & LO, GB & FL           & MS         \\
			ILSTM                      & \cite{guo_exploring_2019}                         & 2019                                                                                                & I    & FI        & LO, GB & ATT          & MS         \\
			KHTT                       & \cite{lin_kernel-based_2022}                      & 2022                                                                                                & I    & FI        & LO, GB & ATT          & MS         \\
			FDT-FTS                    & \cite{silva_c45_2021}                             & 2021                                                                                                & I    & FI        & LO, GB & DT           & MS         \\
			MTNet                      & \cite{chang_memory-network_2018}                  & 2018                                                                                                & I    & FI        & LO, GB     & ATT          & MS         \\
			MAGNN                      & \cite{cheng_financial_2022}                       & 2022                                                                                                & I    & FI        & LO, GB & ATT          & MA         \\
			AT-LSTM                    & \cite{zhang_at-lstm_2019}                         & 2019                                                                                                & I    & FI        & LO, GB & ATT          & MS         \\


			-                          & \cite{pantiskas_interpretable_2020}               & 2020                                                                                               & I    & FI        & LO, GB & ATT          & MS         \\
			Dandelion                  & \cite{zhou_domain_2020}                           & 2020                                                                                               & I    & FI        & LO     & ATT          & MS         \\
			AMS                        & \cite{xu_adaptive_2020}                           & 2020                                                                                               & I    & FI        & LO     & LR           & MS         \\
			IIFI                       & \cite{wang_interpretable_2023}                    & 2023                                                                                               & I    & FI        & GB     & FL           & MS         \\
			CTV-TabNet                 & \cite{seo_exploring_2023}                         & 2023                                                                                               & I    & FI        & LO     & OTH          & MS         \\
			M-MI                       & \cite{zhang_stock_2018}                           & 2018                                                                                               & I    & FI        & GB     & LR           & MS         \\
			~                          & \cite{kim_stock_2018}                             & 2018                                                                                               & I    & FI, VE    & LO, GB & OTH          & MS         \\
			L1-Orth GRU D. Tree        & \cite{wu_stock_2020}                              & 2020                                                                                               & I    & DR        & GB     & APP          & MS         \\
			TABL                       & \cite{tran_temporal_2019}                         & 2019                                                                                               & I    & FI        & GB     & ATT          & MS         \\
			~                          & \cite{yun_interpretable_2023}                     & 2023                                                                                           & I    & FI        & LO, GB & OTH          & MS         \\
			~                          & \cite{feuerriegel_news-based_2019}                & 2019                                                                                           & I    & FI        & LO, GB & LR           & MS         \\
			~                          & \cite{yang_explainable_2019}                      & 2019                                                                                           & I    & FI        & LO, GB & ATT          & MS         \\
			~                          & \cite{wang_interpretable_2020}                    & 2020                                                                                           & I    & FI        & LO     & GRA          & MS         \\
			~                          & \cite{kumar_blended_2021}                         & 2021                                                                                           & I    & FI        & LO     & OTH          & MS         \\
			~                          & \cite{li_one_2021}                                & 2021                                                                                           & I    & DR        & LO, GB & OTH          & MS         \\

			\bottomrule
		\end{tabular}
	}
\end{table*}

\begin{table}[!ht]
  \caption{Definition of the Abbreviation of Characterization Table}
\label{tab:abbreviation}

  \begin{tabular}{cl|cl}
    \toprule
    Acronym & Definition & Acronym & Definition  \\
    \midrule
    E & Explainability &
    I & Interpretability \\
    FI & Feature Importance &
    CL & Confidence Level \\
    VE & Visual Explanation &
    T & Trends \\
    DR & Decision Rule &
    GB & Global \\
    LO & Local &
    APP & Approximation \\
    PER & Perturbation &
    PRO & Propagation \\
    LR & Linear Regression &
    FIN & Financial Theory \\
    DT & Decision Tree &
    ATT & Attention Mechanism \\
    FL & Fuzzy Logic &
    GRA & Graph \\
    OTH & Others &
    MA & Model Agnostic \\
    MS & Model Specific \\

   \bottomrule
  \end{tabular}

\end{table}

\section{Industry Application}
\label{applications}
This section highlights some examples of the application of XAI in finance industries. It presents the results of a poll made in \cite{bhatt_explainable_2020} and proposition for applying XAI in practice.

In industry, XAI is applied in various ways. By surveying industries that use or plan to use XAI, \citeauthor{bhatt_explainable_2020} [\citeyear{bhatt_explainable_2020}] arrived at several interesting findings. Firstly, companies primarily focus on local explainability methods compared to global ones. The needs of companies that drive them to use XAI are debugging, model transparency, model auditing, and monitoring. It is noteworthy that most of these needs are for the internal users of the company and not for the external users. The explainability methods are preferred over interpretable models. A possible reason for this choice is that industries already possess a high-performing model, and they need to understand it in order to enhance it or have it audited. The four types of local explainability that the surveyed industries employ are: feature importance, counterfactual explanations, adversarial training, and influential samples. \citeauthor{bhatt_explainable_2020} [\citeyear{bhatt_explainable_2020}] found that SHAP is the most widely used method among the surveyed industries to showcase the importance of features.

From our end, based on our researches,  we would make the following suggestions for companies that wish to incorporate XAI in their financial time series forecasting works. If a company does not already have a model in use, we suggest creating an interpretable one, as it is also suggested in \cite{leslie_understanding_2019}. However, companies usually have already their own models, so they will need a way to explain it. The model's explainability would serve both internal and external parties. Usually, the first step and the best way to explain a model  is to extract the feature importance allowing to understand which features significantly impact on predictions and which ones do not. These explanations will help refine the model to improve its  performance and allow its decisions to be explained to external users.
On the other hand, a lower number of features can make the explanations easier to understand to the users. However, reducing the number of features may affect the model's performance. A way to solve this problem, without altering the features, is the use of a clustering technique. It is possible to form feature groups with each corresponding to an interpretable concept such as volatility, trend, seasonality, short term, long term, draw down, etc. These concepts are understandable by finance specialists. The importance of each cluster (concept) can be computed by the sum of the importance of the features within the cluster.
Features having a sound financial foundation are preferred because they facilitate linking the model to financial theories. 

When the feature importance is extracted, one can analyze the relationship between time and the importance of various features in making predictions. Local explanations play a crucial role in this analysis. Specifically, they enable creation of samples illustrating how feature importance varies over time periods. These samples can then be examined with respect to specific time frames or regimes. This approach presupposes the ability to categorize time periods into distinct regimes. By characterizing time and relating it with feature importance, the characteristic of the periods can be linked to the features. This analysis would allow users to discern which features are significant at specific times.

\section{Conclusion and Future Works}
\label{conclusion}

In this survey, the most recent interpretable models and explainability methods applied to financial time series prediction are presented. This paper can serve as a useful reference for selecting XAI approaches in finance, as it is specifically composed to assist users in this application. We first outline the VAI problems and provide definitions of some most important concepts established by the community. Then, we introduce two taxonomies 
to guide readers in selecting the appropriate models based on their specific needs. 
We also highlight current industry practices and propose ideas for the effective application of XAI in financial industry.
According to the surveyed publications, feature importance is clearly the preferred XAI principle in financial time series prediction. This preference makes sense because most often users want to know how a model weights its input, like the importance a portfolio manager would naturally give to financial variables. It is also more accessible than the other principles. 

 There appears to be a greater focus on developing interpretable models as compared to explainable models. This trend could be due to the context dependancy of the financial models. 
 Indeed, models used for financial time series prediction are often specifically designed for this task. The designers of these models might want to develop models that could be understandable by humans, namely XAI models, to apply them in practice. It seems logical to create a model with some interpretable components from the beginning, rather than creating a black-box model and explaining it later. In fact, very often, the concepts and the techniques of explainability methods could be directly applied to AI models to make them interpretable or to improve their interpretability and accuracy. Nothing prevents explaining the interpretable model later using an explainability method if the interpretations are insufficient.

 Based on this review, we can say that linear regression, attention mechanism and ruled based models are the most popular techniques for making an interpretable model, while the SHAP algorithm is the most prevalent among explainability methods. The popularity of SHAP is due to its robust mathematical foundation and ease of use.

We emphasize that explainability and interpretability are distinct concepts that should be treated separately. This distinction is necessary because of their unique applications and utilization. According to \cite{rudin_stop_2019, leslie_understanding_2019}, interpretable models should be preferred to explainability methods applied to black-box models in high-risk decision-making. 
For instance, the authors of \cite{rudin_stop_2019} strongly argue against the necessity of a trade-off between accuracy and interpretability in model design supported by examples in  \cite{pantiskas_interpretable_2020, chen_holistic_2022}. Also, explanations from an explainability method are not always faithful to the model and are not always accurate or complete. In contrast, the interpretable principles of a model are, by definition, faithful to it. Moreover, unlike 
 interpretable models, it is difficult to add external data to a black-box model because we do not know what information the model uses and what needs to be added.

The interpretable models are not perfect. There are problems related to interpretable models too, according to \cite{rudin_stop_2019}. Indeed, there is a computational challenge in training these models due to the complex architecture, made of many different parts, that interpretable models must have. Also, development of effective interpretable models often needs prior knowledge as compared to a black-box model that can sometimes learn surprising patterns. These patterns can be learned by an interpretable model, but this model must be specifically designed. It usually takes more time to develop an interpretable model than a black box one to obtain the same accuracy. Additionally, the reliability of the interpretation of complex models seems to be questionable as discussed in Subsection \ref{inter_eval}. Also, in scenarios where a well-performing model is already in use, it may be reasonable to apply some explainability methods to decipher the model than to rebuild new interpretable ones.

We would like to finish the paper by some final remarks. 
Few studies, if any, rigorously tested the faithfulness and the reliability of their explainability methods; a ranking based on explainability levels could be enlightening.
Moreover, the absence of a quantitative metric for model interpretability suggests the need for further work in this domain.
Additionally, since most XAI models have not been tested with the same benchmarks regarding the performance and the interpretability/explainability, the selection of the best approach remains challenging. More quantitative studies will be necessary for the development and the application of the XAI approaches.

\begin{acks}
     We would like to thank Patrick Asante Owusu and Simon Lévesque for their comments on the article. 
\end{acks}

\bibliographystyle{ACM-Reference-Format}
\bibliography{biblio}

\end{document}